\documentclass[a4paper]{cas-sc}

\usepackage[numbers]{natbib}
\usepackage{graphicx}
\usepackage{float}
\usepackage{hyperref}

\def\tsc#1{\csdef{#1}{\textsc{\lowercase{#1}}\xspace}}
\tsc{WGM}
\tsc{QE}

\newcommand{\R}{\mathbb{R}}

\begin{document}
\let\WriteBookmarks\relax
\def\floatpagepagefraction{1}
\def\textpagefraction{.001}

\shorttitle{Neural Networks for Threshold Dynamics Reconstruction}    

\shortauthors{E. Negrini, A. J. Gao, A. Bowering, W. Zhu, L. Capogna}  

\title [mode = title]{Neural Networks for Threshold Dynamics Reconstruction} 
\tnotemark[1]
\tnotetext[1]{\textbf{Funding:} E.N. is partially funded by NSF DMS-2331033. W.Z. is partially supported by NSF DMS-2244976 and DMS-2502900. L.C. is partially funded by  a Smith College CEEDS Fellowship, and by the awards NSF DMS-195599 and DMS-2348806. A.B. and A.G. are partially funded by  a Smith College CEEDS Fellowship.}



%

\author[1]{Elisa Negrini}[orcid=0000-0001-6647-0046]

\cormark[1]


\ead{enegrini@math.ucla.edu}


\credit{Conceptualization, Methodology, Software}

\affiliation[1]{organization={Department of Mathematics, University of California Los Angeles}, 
            city={Los Angeles},
            state={CA},
            country={USA}}

\author[2]{Almanzo Jiahe Gao}


\ead{agao@smith.edu}


\credit{Real Data Preprocessing}

\affiliation[2]{organization={Department of Mathematical Sciences, Smith College},
            city={Northampton},
            state={MA},
            country={USA}}

\author[3]{Abigail Bowering}


\ead{bowering@mit.edu}


\credit{Real Data Preprocessing}

\affiliation[3]{organization={Institute for Data, Systems, and Society, Massachusetts Institute of Technology},
            city={Cambridge},
            state={MA},
            country={USA}}

\author[4]{Wei Zhu}[orcid=0000-0002-9181-5103]


\ead{weizhu@gatech.edu}


\credit{Conceptualization, Methodology}

\affiliation[4]{organization={School of Mathematics, Georgia Institute of Technology},
            city={Atlanta},
            state={GA},
            country={USA}}

\author[2]{Luca Capogna}[orcid=0000-0002-1839-8462]


\ead{lcapogna@smith.edu}


\credit{Conceptualization, Methodology}


\cortext[1]{Corresponding author}



\begin{abstract}
We introduce two convolutional neural network (CNN) architectures, inspired by the Merriman-Bence-Osher (MBO) algorithm and by  cellular automatons, to model and learn threshold dynamics for front evolution from video data. The first model, termed the (single-dynamics) MBO network, learns a specific kernel and threshold for each input video without adapting to new dynamics, while the second, a meta-learning MBO network, generalizes across diverse threshold dynamics by adapting its parameters per input. Both models are evaluated on synthetic and real-world videos (ice melting and fire front propagation), with performance metrics indicating effective reconstruction and extrapolation of evolving boundaries, even under noisy conditions. 
Empirical results highlight the robustness of both networks across varied synthetic and real-world dynamics.
\end{abstract}




\maketitle

\section{Introduction}
In 1992 Merriman, Bence, and Osher formulated in \cite{MBO,MBO1} the so called Merriman-Bence-Osher (MBO) algorithm. This algorithm provides a computational method to track the time evolution of a set whose boundary moves with a normal velocity equal to a dimensional constant times its mean curvature, i.e., {\it motion by mean curvature}. The algorithm is  based on  a  time discrete thresholding scheme in which a linear operation (heat diffusion) and a nonlinear operation (thresholding) are applied iteratively. By {\it thresholding} a function $f$, with a given threshold $a$, we indicate the operation of  substituting $f$ with the characteristic function of its super level set $\{f \ge a\}$.  More specifically, if one denotes by
$H(s)$ the Heaviside function
$$H(s)= \begin{cases}1 & \text{ if } s\ge 0\\
0 & \text{ if } s< 0\end{cases},$$ then starting from an connected open set $\Omega \subset \mathbb R^n$, and setting $u^0=\chi_{\Omega}:\R^n\to \R$ to be its characteristic function, the MBO algorithm produces a sequence of  functions $\{u_h^{N}\}_{N\ge 0}$ given by
\begin{align}
\label{eq:mbo_original}
    u_h^0 = u^0, \quad u_h^{N+1}:= H \left[ \left(K_h* u_h^N\right)-\frac{1}{2}\right],
\end{align}
where $K_h(x) = (4\pi h)^{-n/2}\exp\left( -\frac{|x|^2}{4h}\right)$ denotes the heat kernel at time $h>0$, with $h$ as a fixed parameter, and $K_h* u_h^N$ is the \textit{convolution} between $K_h$ and  $u_h^N$, given by
\begin{align}
\label{eq:convolution}
    K_h* u_h^N(x) =\int_{\R^n} K_h(y) u_h^N(x-y) dy.
\end{align}
The Heaviside function in Eq.~\eqref{eq:mbo_original} corresponds to applying a thresholding operation with $a=\frac{1}{2}$, and the resulting thresholded functions, $u^N_h$, are characteristic functions of sets $C_h^N$. 
Evans~\cite{E} proved that for any $t>0$ ,
$$C_{\frac{t}{m}}^m \to M(t) C^0 \text{ as } m\to \infty,$$
where $M(t)C^0$ represents the level set $\left\{x \in \mathbb{R}^n \mid u(x, t) = 0\right\}$ at time $t > 0$ of the viscosity solution $u(x, t)$ to the generalized mean curvature flow of $C^0$, given by
\begin{equation}\label{MCF}
\partial_t u = \sum_{i,j=1}^n \bigg(\delta_{ij} - \frac{u_i u_j}{|\nabla u|^2} \bigg) u_{ij} \end{equation}
(see  \cite{CGG, ES} for the pertinent definitions and \cite{BG,Ma,IPS} for other proofs of this convergence). 

\medskip

The paper \cite{IPS} provides proofs of convergence for more general thresholding evolutions of fronts, including the \textit{cellular automaton} models described in \cite{GrGr}. As outlined in \cite[(0.2)-(0.4)]{IPS}, these models describe the evolution of a front through the following scheme: Let $\mathcal N\subset \mathbb R^n$ denote an open bounded neighborhood of the origin with unit measure. Starting from a configuration $A\subset \R^n$ and fixing parameters $h>0$ and $0<\theta<1$, one sets 
\begin{align}
\label{eq:cellular}
    M_h (A) = \left\{x\in \mathbb R^n \text{ such that } |(x+h\mathcal N ) \cap A|\ge \theta h^n\right\}.
\end{align}
In other words, quoting \cite{IPS}: {\it ``If $A$ is the occupied set at time $t$, the occupied set $M_h(A)$ at time $t+h$ consists of those points for which the volume of the overlap between $x+h\mathcal N$ and $A$ exceeds the quantity $\theta|h \mathcal N|$''}. Observe that the measure of the set $(x+h\mathcal N ) \cap A$ can also be expressed as the \textit{convolution} of the characteristic function of $A$ with the characteristic function of $-h\mathcal N$, evaluated at $x\in \R^n$, i.e.,
\begin{align}
    \left|(x+h\mathcal N ) \cap A\right| = \chi_{A} * \chi_{-h\mathcal{N}}(x).
\end{align}

We also note that in \cite{MBO} the authors describe how, changing the kernel and adjusting the threshold allows the MBO algorithm to effectively approximate the dynamics of various types of evolving fronts. For instance, by choosing a kernel $K$ as the characteristic function of the unit ball and setting the threshold to zero (instead of $1/2$ in Eq.~\eqref{eq:mbo_original}), one recovers the motion of a front moving along its normal at constant speed (as described in the flame propagation model in \cite{Se, Ba}). 

Since both the MBO algorithm~\eqref{eq:mbo_original} and the cellular automaton model~\eqref{eq:cellular} describe the evolution of fronts through alternating convolutions with a kernel and thresholding, we will refer to the evolutions of sets by either algorithm as {\it thresholding scheme evolutions} or {\it threshold dynamics} throughout this paper. Both models, along with their variants, will be included in our analysis.

\subsection{Our contribution}
The alternation of a linear operator (convolution with a kernel) and a nonlinear operator (thresholding implemented through the Heaviside function) in the MBO algorithm and the cellular automaton process is reminiscent of the similar interplay between linear and nonlinear operations in artificial neural networks. In this paper we pursue this analogy and implement variants of the MBO algorithm and various cellular automaton models using  convolution  neural networks (CNNs). Our primary motivation is to solve the following inverse problem:

\medskip

\noindent {\bf Inverse Problem}: {\it Given $N$ consecutive frames in a video depicting an evolving front, determine the kernel and threshold corresponding to a threshold dynamics that best approximates the observed evolution.}

\medskip

We address this problem as an {\it unsupervised learning} task and propose two different but related approaches: 

\begin{itemize}
    \item In the first approach, the thresholding algorithm is implemented by associating each time step with a layer in a recurrent CNN, which we refer to as the {\bf MBO network}. In this architecture, all layers of the  network share \textit{a single kernel} and \textit{a single threshold}, both of which are treated as trainable weights and learned during training. For example, given a video consisting of four frames, the input to the MBO network is the first frame of a video, and the network outputs three predicted frames that represent the subsequent evolution according to the MBO algorithm.
    
    To solve the inverse problem, we use the mean squared error (MSE) between the predicted frames and the actual frames in the video as the loss function. Minimizing this loss function enables the network to learn the kernel and threshold that best fit the observed evolution. This method, with its pros and cons, is described more in detail in Section \ref{2.1}. 
 
 \item  In the second approach, we propose a \textbf{meta-learning MBO network}, which functions as a hypernetwork designed to adapt the kernels and thresholds of an underlying MBO network to accommodate diverse, potentially unseen evolutions. Specifically, consider a video of four frames depicting a moving front. The meta-learning MBO network takes the entire set of frames, $I_1, \dots, I_4$, as input and outputs a kernel and a threshold. These outputs are then transferred as weights to the MBO network, which predicts the subsequent evolution $\hat{I}_2, \hat{I}_3, \hat{I}_4$ starting from the first frame $I_1$. 

The loss function measures the discrepancy between the predicted frames $\hat{I}_j$ and the original frames $I_j$ across a diverse dataset of videos featuring various types of front evolutions. This approach is detailed in Section \ref{2.2}. Its key advantage is its ability, once properly trained on videos of diverse thresholding scheme evolutions, to generalize and infer the kernel and threshold parameters for an arbitrary thresholding scheme evolution.
\end{itemize}

We train and test these two models on both synthetic and real data. The synthetic data consists of videos generated from thresholding scheme evolutions using a variety of kernels, ranging from Gaussian kernels to the grayscale functions corresponding to MNIST digits. The real data includes videos of ice melting (which roughly corresponds to mean curvature flow) and forest fires (which corresponds to unit normal velocity of the front). To assess the models' robustness under noisy conditions, we solve the inverse problems for these videos with noise (Gaussian blur and salt-and-pepper noise) as well as in noise-free scenarios. See Section \ref{real_data} for more details.

The accuracy of the models is evaluated using three metrics: mean squared error (MSE), structural similarity index (SSIM), and Jaccard index. Since the size of the kernel associated with a given thresholding scheme evolution video is not known a priori, we also test the models for robustness with respect to varying kernel dimensions.

The code and data for this work are available at \url{https://github.com/enegrini/MBO_network.git}


\subsection{Related work}

Originally proposed as geometric front evolution models, threshold dynamics have found applications in data science for partitioning and clustering problems. In computer vision, threshold dynamics have been employed in tasks like image segmentation~\cite{ESEDOGLU2006367} and shape reconstruction~\cite{ERT}. The MBO scheme, in particular, has been adapted for graph-based approaches to solve semi-supervised data clustering problems~\cite{bertozzi2012diffuse, garcia2014multiclass, merkurjev2014graph}. Variants of MBO schemes, such as the Volume-Constrained MBO~\cite{jacobs2018auction} and Poisson MBO~\cite{calder2020poisson}, have further extended its utility to clustering with volume constraints or at very low label rates. Recent theoretical analyses have examined the large-data limit of the MBO scheme for clustering~\cite{JMLR:v24:22-1089, handle:20.500.11811/10889}. Despite the extensive literature on applying threshold dynamics to machine learning tasks, to the best of our knowledge, no studies address the inverse problem: identifying (unknown) threshold dynamics from available data.

\section{Methodology}
We propose two methods that combine convolutional neural networks (CNNs) and Merriman-Bence-Osher (MBO) scheme to model and learn threshold dynamics for front evolution from video data.

\subsection{Method 1: MBO network}\label{2.1}
The first method directly implements the MBO scheme by successively applying a convolutional layer followed by thresholding at each layer.  All layers share the same kernel and threshold, both of which are learned during training.
Since the backpropagation and gradient descent methods that are used in the training of the neural network involve derivatives of the network with respect to the parameters, in our setting we need to substitute the Heaviside thresholding function with a smooth analogue, the sigmoid $$\sigma(x)= \frac{1}{1+e^{-x}}.$$
We will also modify this function according to two parameters $a=\text{threshold} \in (0,1)$ and $s=\text{steepness} >0$,
setting
\begin{equation}\label{steepness}
    \sigma_{s,a}(x) = \frac{1}{1+\exp(-s(x-a))}
\end{equation}
In the case of Gaussian kernel and threshold equal to $1/2$, this substitution gives rise to a new thresholding scheme defined as
\begin{equation}\label{new_scheme}
    \tilde u_h^{N+1}:= \sigma_{s,a}\bigg[K_h*  \tilde u_h^N\bigg].
\end{equation}
During testing, the Heaviside function is reinstated. This architecture, essentially a recurrent CNN with a specialized nonlinearity, is referred to as the \textbf{MBO network}.

\begin{figure}[h!]
    \centering
    \includegraphics[width=\linewidth]{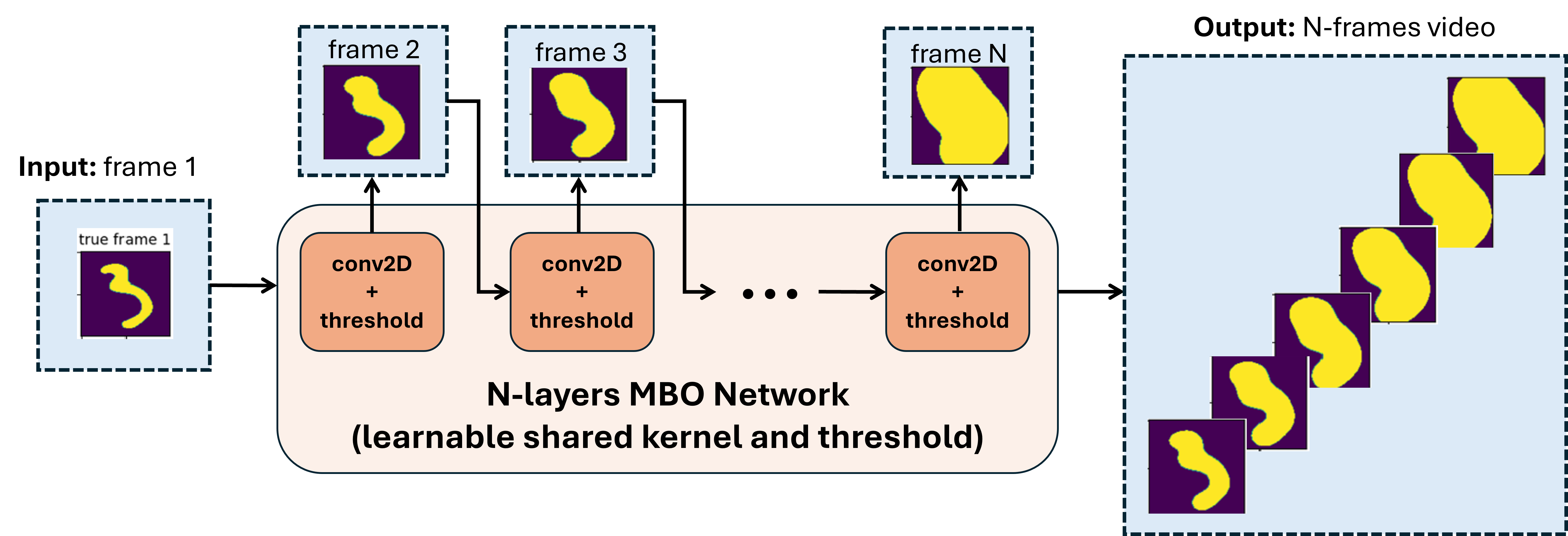}
    \caption{MBO network architecture}
    \label{fig:MBOArch}
\end{figure}

In this approach, the model learns a single kernel and threshold for each video, meaning that a unique MBO network must be trained for each individual video. While this method is capable of accurately capturing the dynamics for a specific video, its generalization capacity is limited, since the kernel and threshold are tailored only to the given video sequence. A representation of this architecture is shown in Figure \ref{fig:MBOArch}.

\noindent
\textbf{Training and Testing Procedure}\\
Assume, for example, that we are given a collection of $K$ short videos, each consisting of four frames and sharing the same underlying threshold dynamics (i.e., the same threshold and kernel). During training, the network receives as input the first frame of a video, and the target consists of the next three consecutive frames. The network itself is composed of three layers, each sharing the same learnable kernel and threshold. Each layer of the network produces one frame as output, and the three frames generated by the three layers are then combined to form a predicted video. The loss is computed by comparing these generated frames with the actual frames from the video using mean squared error (MSE). Specifically, for video $j$, the input frame is \( I_1^j \), the network predicts frames \( \hat{I}_2^j, \hat{I}_3^j, \hat{I}_4^j \), and the true frames are \( I_2^j, I_3^j, I_4^j \). The total loss is then computed as:
\begin{align*}
\text{Loss} =\frac{1}{K} \sum_{j=1}^{K} \left( \frac{1}{M} \sum_{i=2}^{4} \left\| I_i^j - \hat{I}_i^j \right\|_2^2 \right)
\end{align*}
where \( K \) is the number of training videos, and \( M \) is the number of pixels per frame. The network learns by updating the kernel and threshold to minimize this loss. It is worth emphasizing that the training is entirely unsupervised: the true kernel and threshold are not known or used in the loss function; only frames \( \hat{I}_2^j, \hat{I}_3^j, \hat{I}_4^j \) are used. For our loss function we decided to use MSE as we observed a fast convergence during training. However, as explained in Section~\ref{metrics}, other metrics may be more appropriate than MSE to capture image discrepancies. The exploration of alternative, problem-specific metrics is deferred to future work.

During testing, the learned kernel and threshold remain fixed. The first frame of a new video (sharing the same dynamics as the training examples) is provided as input, and the trained MBO network generates a video of length $N$. Notably, this approach allows for training and testing on videos of different lengths, where $N$ can exceed the number of layers used during training (e.g., three layers). In our experiments, we set $N = 7$  to evaluate whether the learned kernel and threshold can effectively be used to extrapolate future frames.

\noindent\textbf{Pros and Cons}\\
The MBO network has the advantage of directly learning dynamics specific to the input video, allowing it to accurately capture the governing law of the observed dynamics. This approach is particularly effective when applied to a set of videos likely to share similar dynamics, as it enables fast training and can perform well even with a short, four-frame video (see Section \ref{sec:one_vid}). Furthermore, the model’s simplicity---using a single kernel and threshold across layers---reduces the number of parameters, resulting in a compact and efficient network that is easier to optimize.

This method is especially useful when prior knowledge suggests that the same kernel and threshold can be shared across multiple videos, leveraging fast training while maintaining good predictive performance in such scenarios. However, if multiple videos with different underlying dynamics (i.e., kernels and thresholds) are used for training, the learned kernel and threshold may converge to an ``average'' of the true values for the individual videos. This averaging effect can compromise the network's ability to accurately predict the dynamics of any specific video, as it loses the capacity to capture the unique characteristics of each video's evolution. Our second method addresses this limitation.

\subsection{Method 2: Meta-learning MBO network}\label{2.2}
To overcome the limitations of the first method and improve generalization, we propose the \textbf{meta-learning MBO network}, an architecture that combines a trainable convolutional neural network with an MBO network with frozen weights. In this approach, the trainable CNN aims to produce kernels and thresholds based on the input video. These generated kernels and thresholds are then frozen and used by the subsequent MBO network to produce the video prediction. This effectively forms a hypernetwork, where the trainable CNN determines the weights (kernels and thresholds) of the MBO network.

\begin{figure}[h!]
    \centering
    \includegraphics[width=\linewidth]{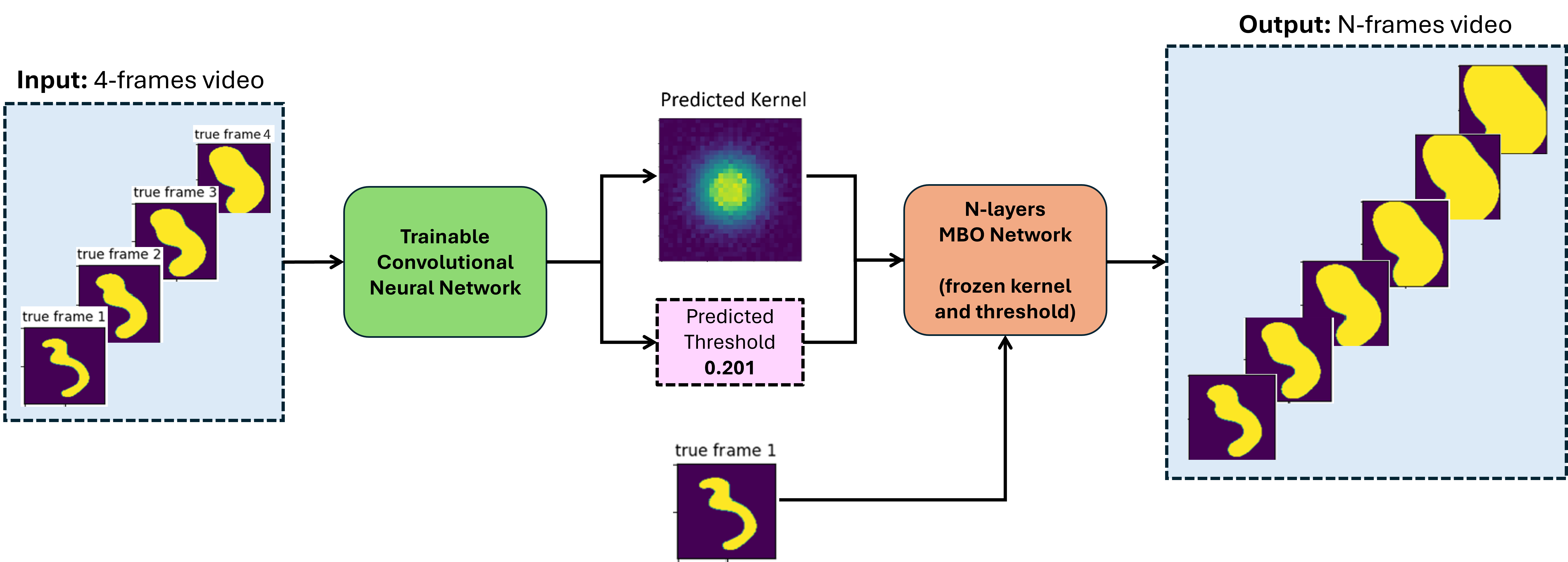}
    \caption{Meta-learning MBO network architecture}
    \label{fig:MetaLearn}
\end{figure}

This method provides significant flexibility, as the convolutional network can adaptively generate unique kernels and thresholds for each input video. It enables the use of a single network to process a variety of input data without requiring retraining for each new video. A representation of the meta-learning MBO architecture can be found in Figure~\ref{fig:MetaLearn}.

\noindent
\textbf{Training and Testing Procedure}\\
In the training phase, the input consists of a four-frame video (instead of only the first frame in Method~1). This video is passed through the trainable CNN, which produces the kernel and threshold corresponding to the video’s dynamics. These parameters are then frozen and passed to the subsequent MBO network, which, along with the first frame, generates frames 2, 3, and 4. The MBO network in this setup has three layers, each sharing the same kernel and threshold that were produced by the trainable CNN. Again, to avoid singularities in backpropagation, during training, we use a smoothed thresholding through a sigmoid function as in \eqref{steepness}.

The loss is computed by comparing the predicted frames 2, 3, and 4 with the corresponding true frames from the input video, enabling the network to optimize the kernel and threshold generation in the CNN block. Specifically, for a video $j$ given with input frames \(I_1^j, I_2^j, I_3^j, I_4^j\), the meta-learning MBO network predicts frames \(\hat{I}_2^j, \hat{I}_3^j, \hat{I}_4^j\), and the loss is calculated as:
$$
\text{Loss} =\frac{1}{K} \sum_{j=1}^{K} \left( \frac{1}{M} \sum_{i=2}^{4} \left\| I_i^j - \hat{I}_i^j \right\|_2^2 \right),
$$
where \( K \) is the total number of training videos and \( M \) is the number of pixels per frame. This loss trains the CNN block to effectively generate the kernel and threshold for each training video sequence $I^j$.

During testing, the trained CNN block produces a kernel and threshold based on a new four-frame video, potentially with a \textit{different} underlying dynamics. These parameters are then passed to the MBO network, which is now extended to $N$ layers (where $N$ can be greater than 3) to generate an output video of $N$ frames. This setup allows the network to predict frames further into the future. During testing, the Heaviside function is reinstated in thresholding step. In our experiments, we set $N = 7$ to evaluate whether the learned kernel and threshold could effectively generalize and extrapolate additional frames beyond the three frames used during training.\\
\textbf{Pros and Cons}\\
The meta-learning MBO network offers several notable advantages. First, it enables the use of a single network to generate videos across a wide variety of inputs that may come from different domains or possess different threshold dynamics (i.e. having different kernels and thresholds). This provides a significantly higher level of generalization and adaptability compared to the first method, as the CNN can learn to generate different kernel-threshold combinations tailored to each video. Furthermore, once trained, the model can be applied to new videos without requiring retraining, making it an efficient and flexible solution for applications involving large datasets or diverse video dynamics.

The main drawback of this technique is its increased complexity, which demands more training time compared to the first method. The meta-learning approach involves training not only the convolutional network but also optimizing the entire system to ensure that the generated kernel and threshold work effectively within the  MBO framework. This can result in longer training times and requires more hyperparameter tuning.

\subsection{Comparison of the Two Methods}
The MBO network (Method 1) is well-suited for scenarios where training time is limited, and the videos are relatively homogeneous. It allows for quick and efficient training for a small group of similar videos that share the same underlying dynamics. However, it lacks the flexibility needed to generalize across multiple videos with differing dynamics, making it less suitable for diverse datasets.

On the other hand, the meta-learning MBO network (Method 2) is ideal for applications involving diverse datasets with varying dynamics. Its ability to adapt flexibly to different inputs without requiring retraining offers superior generalization and adaptability. However, this comes at the expense of longer training times due to the increased complexity of the network. This approach is particularly advantageous when the objective is to handle videos with different underlying kernels and thresholds, especially when the model needs to be applied across a wide variety of inputs.

\subsection{Comparison Metrics}\label{metrics}
To evaluate the accuracy of the reconstructed video sequence, we use three metrics: Structural Similarity Index (SSIM), Jaccard Index (Jac. idx), and Relative Mean Squared Error (Relative MSE).

SSIM quantifies the structural similarity between two images, taking into account luminance, contrast, and structure. SSIM provides a value between -1 and 1, where 1 indicates perfect similarity. For two images $x$ and $y$, their SSIM is defined as:
    $$\text{SSIM}(x, y) = \frac{(2\mu_x \mu_y + C_1)(2\sigma_{xy} + C_2)}{(\mu_x^2 + \mu_y^2 + C_1)(\sigma_x^2 + \sigma_y^2 + C_2)},$$
where $\mu_x$ and $\mu_y$ are the means of $x$ and $y$, $\sigma_x^2$ and $\sigma_y^2$ are their variances, $\sigma_{xy}$ is the covariance between $x$ and $y$, and $C_1,C_2>0$ are constants to avoid division by zero. In our experiments, we compute the SSIM for each frame and average the values across all frames in a video to obtain a video-level SSIM.

The Jaccard Index (Jac. idx), also known as Intersection over Union (IoU), measures the similarity between two sets by calculating the ratio of the intersection to the union. In the context of binary images, it quantifies the overlap between foreground pixels, giving a value between 0 (no overlap) and 1 (perfect overlap). For binary images $x$ and $y$, the Jaccard Index is defined as:
    $$\text{Jaccard}(x, y) = \frac{|x \cap y|}{|x \cup y|}.$$
As with SSIM, we compute the Jaccard Index for each frame and average across all frames in a video to obtain the video-level Jaccard Index.

Finally, the Relative Mean Squared Error (Relative MSE) measures the difference between the predicted video and the true video, normalized by the true video. For predicted video frames $x_i$ and ground truth frames $y_i$, the Relative MSE is given by:
    $$\text{Relative MSE}(x, y) = \frac{\sum_{i=1}^{N} \| x_i - y_i \|^2}{\sum_{i=1}^{N} \| y_i \|^2 +\epsilon} ,$$
where $N$ is the number of video frames, $\|\cdot\|^2$ denotes the squared Frobenius norm, and $\epsilon$ is a small number added for stability. Note that Relative MSE may not be the best metric for comparing binary images, especially in scenarios where the true frame is empty or has very few non-zero pixels (this happens often for thresholds greater than 0.5 where the subject shrinks in time). In such cases, the MSE can be disproportionately influenced by the presence of a few non-zero values, leading to misleading interpretations of similarity. This is because a small change in a sparse area can yield a high MSE despite the overall structural similarity being maintained.

For each test video sequence, we compute SSIM, Jaccard Index, and Relative MSE by comparing the true test video with the reconstructed video generated using the learned kernel and threshold. We then average these metrics across all test videos to evaluate overall performance.

\section{Data generation}
\subsection{Synthetic data}\label{synt_data}
The synthetic dataset is generated using variants of the MBO scheme with different kernels and thresholds, as described in the introduction. Starting with an MNIST digit as the initial frame, we iteratively apply a convolution with a kernel followed by thresholding using the Heaviside function. This process generates a video sequence of threshold dynamics over \( N \) frames, where \( N \) corresponds to the number of convolution-thresholding operations performed.

\begin{figure}[h!]
    \centering
    \includegraphics[width=0.8\linewidth]{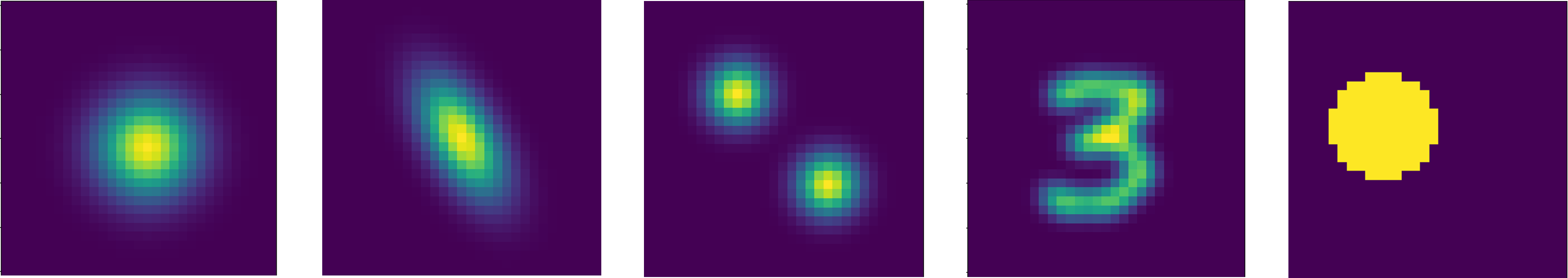}
    \caption{Example of kernels used for dataset generation. From left to right: standard Gaussian kernel, skewed Gaussian kernel, double Gaussian kernel, MNIST digit, and the indicator function of a disk.}
    \label{fig:kerEx}
\end{figure}

In order to increase the variability of our dataset, we generate data using a variety of kernels, including variants of the Gaussian kernel, indicator functions of disk, and MNIST digits. A general 2D Gaussian kernel is defined as:
\begin{equation}\label{Gaussian}
    K(x, y) = \frac{1}{2\pi\sigma_x\sigma_y} \exp \left( -\frac{(x - \mu_x)^2}{2\sigma_x^2} - \frac{(y - \mu_y)^2}{2\sigma_y^2} \right),
\end{equation}
where \( \mu_x \) and \( \mu_y \) are the means, and \( \sigma_x \) and \( \sigma_y \) are the standard deviations along the \( x \)- and \( y \)-axes, respectively. The types of kernels used in our experiments include:
\begin{enumerate}
    \item Standard Gaussian kernel, where \( \mu_x = \mu_y \) and \( \sigma_x = \sigma_y \), resulting in a symmetric Gaussian spread.
    
    \item Skewed Gaussian kernel, where \( \sigma_x \neq \sigma_y \), introducing asymmetry into the kernel shape.
    
    \item Double Gaussian kernel, a mixture of two Gaussian distributions centered at different locations, possibly with different variances.
    
    \item A randomly picked MNIST digit.

    \item Indicator function of a disk with variable center and radius.
\end{enumerate}
Examples of these kernels are shown in Figure~\ref{fig:kerEx}.

\begin{figure}[h!]
    \centering
    \includegraphics[width=\linewidth]{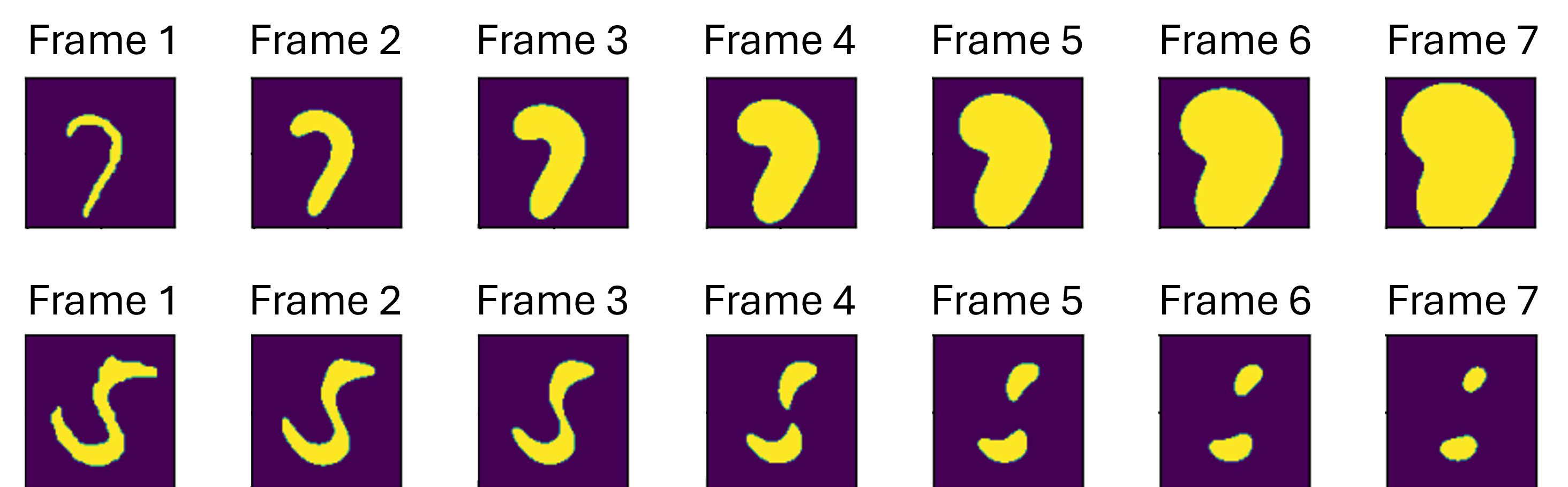}
    \caption{Example of dynamics obtained using the MBO scheme with standard Gaussian kernel and threshold 0.2 (top) and 0.5 (bottom).}
    \label{fig:thrEx}
\end{figure}

For the thresholds, we use values ranging from 0.2 to 0.7. When the threshold is below 0.5, the object in the video expands over time, while thresholds above 0.5 cause the object to shrink as the dynamics progress. This combination of kernels and thresholds produces a diverse range of behaviors in the generated dataset. An example of the dynamics produced with a standard Gaussian kernel using thresholds of 0.2 and 0.5 is shown in Figure \ref{fig:thrEx}.

\begin{figure}[h!]
    \centering
    \includegraphics[width=\linewidth]{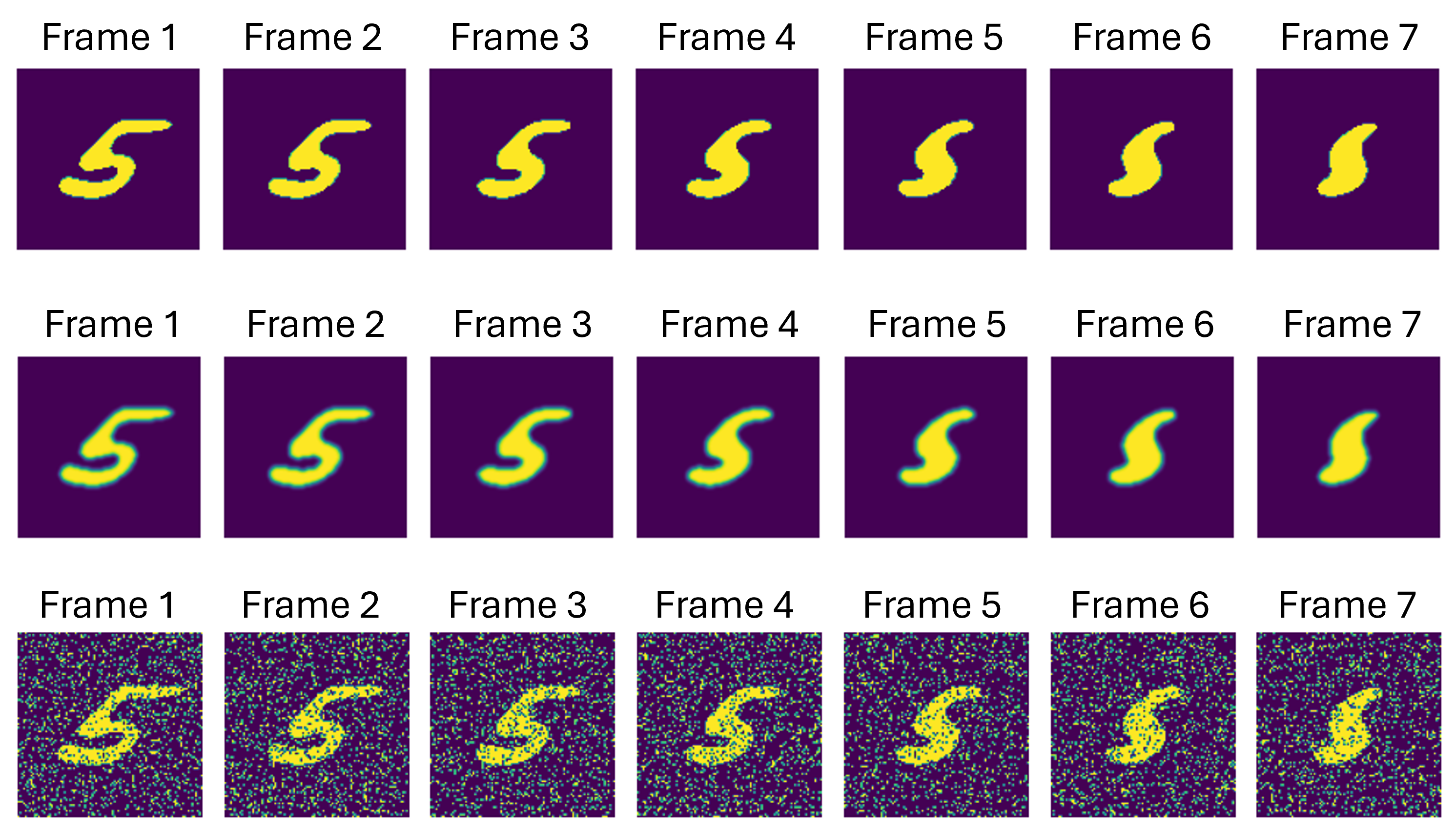}
    \caption{Example noise conditions on threshold dynamics generated with standard Gaussian kernel and threshold 0.5. \textbf{Top:} No noise. \textbf{Middle:} Gaussian blur. \textbf{Bottom:} Salt-and-pepper noise.}
    \label{fig:noiseEx}
\end{figure}

Finally, in our experiments, we generate video sequences under three different noise conditions: without noise, with Gaussian blur, and with salt-and-pepper noise. Gaussian blur is applied by convolving the image with a Gaussian kernel, defined as:
$$
I_{\text{blurred}}(x, y) = \sum_{i, j} I(x-i, y-j) \cdot K(i, j),
$$
where \( I(x, y) \) is the original image and \( K(i, j) \) is the 2D standard Gaussian kernel. Salt-and-pepper noise is defined by randomly setting a fraction of the pixels to either 0 (black) or 1 (white), introducing high-contrast noise into the image. Specifically, for each pixel \( I(x, y) \), with probability \( p_{\text{noise}} = 0.3 \), the pixel value is replaced by either 0 or 1. An example of the dynamics under the different noise conditions can be seen in Figure \ref{fig:noiseEx}.

We observe that salt-and-pepper noise is the most challenging to handle, as the abrupt, random changes in pixel values make it difficult for the method to identify meaningful structures in the image. In contrast, Gaussian blur poses less difficulty, as the smoothing effect of the blur aligns with the behavior of the sigmoid thresholding used during training, enabling the method to adapt and effectively learn the underlying dynamics.

\subsection{Real data}\label{real_data}
This section outlines the data collection and preprocessing methods for real-world data, specifically focusing on fire front and ice-melting dynamics.

\begin{figure}[h]
    \centering
    \includegraphics[width=0.8\linewidth]{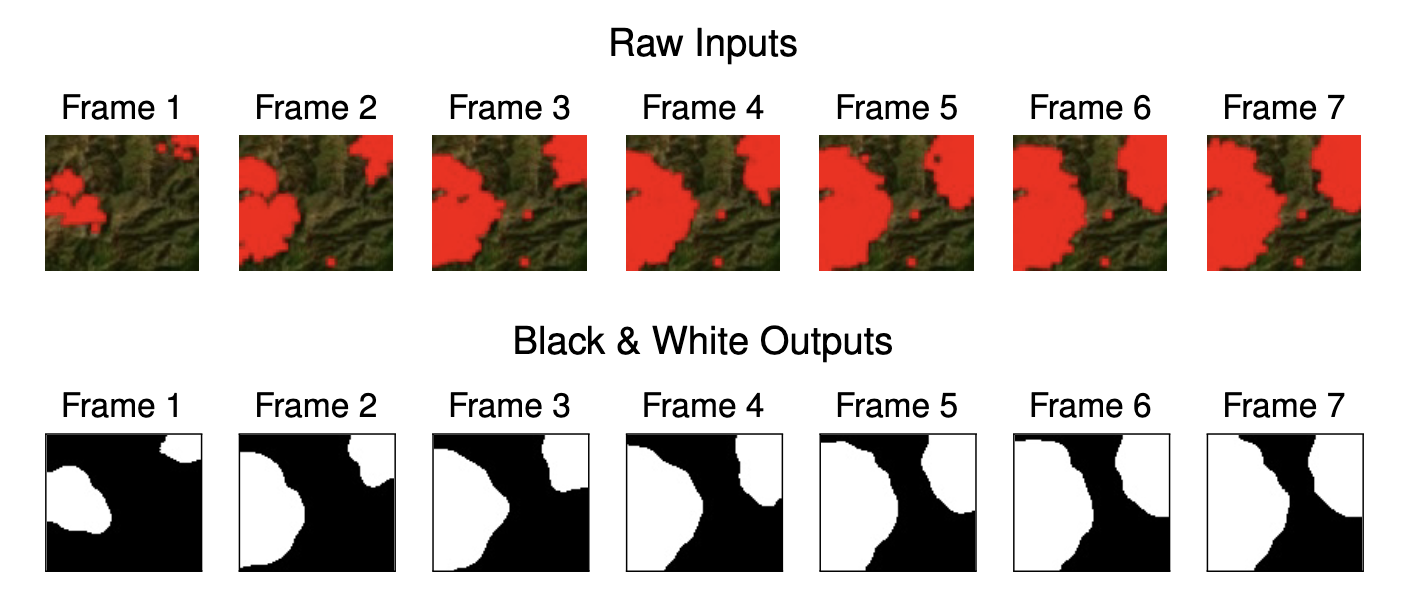}
    \caption{Example of one fire front expansion video. \textbf{Top: } Raw inputs. \textbf{Bottom: }Processed data.}
    \label{fig:fires}
\end{figure}

\noindent \textbf{Fire fronts dataset.}
The initial data was collected from NASA-FIRMS (NASA-Fire Information for Resource Management System). This dataset provides interactive browsing of 
\textit{"the full archive of global active fire detections from MODIS and VIIRS. Near real-time fire data are available within approximately 3 hours of satellite overpass and imagery within 4-5 hours,"} \cite{firms}. When recording images from NASA-FIRMS, we focus on sections where the fire boundary expands approximately uniformly in order to optimize performance in the MBO algorithm. In fact, natural fire expansion is often influenced by external factors such as wind and geographical obstacles like rivers, leading to non-uniform growth. The datasets are collected such that they represent the cumulative burnt area at each daily time step. For example, the second frame in Figure \ref{fig:fires} represents the combined area from the first and second day of the fire. The sequence of images is stopped when there is no significant fire expansion across its boundary.


The images are pre-processed in Python to extract seven suitable images capturing the fire front dynamics. First, as the NASA-FIRMS data are pixelated at a resolution of 375m, we apply a Gaussian blur to smooth boundaries. The images are then converted into HSV color space to more easily distinguish the active fire regions. Then, we convert the images to binary black-and-white images. The resulting images are used as input data for the MBO networks. Figure \ref{fig:fires} provides an example of the raw and processed fire front expansion data.

\begin{figure}[h]
    \centering
    \includegraphics[width=0.8\linewidth]{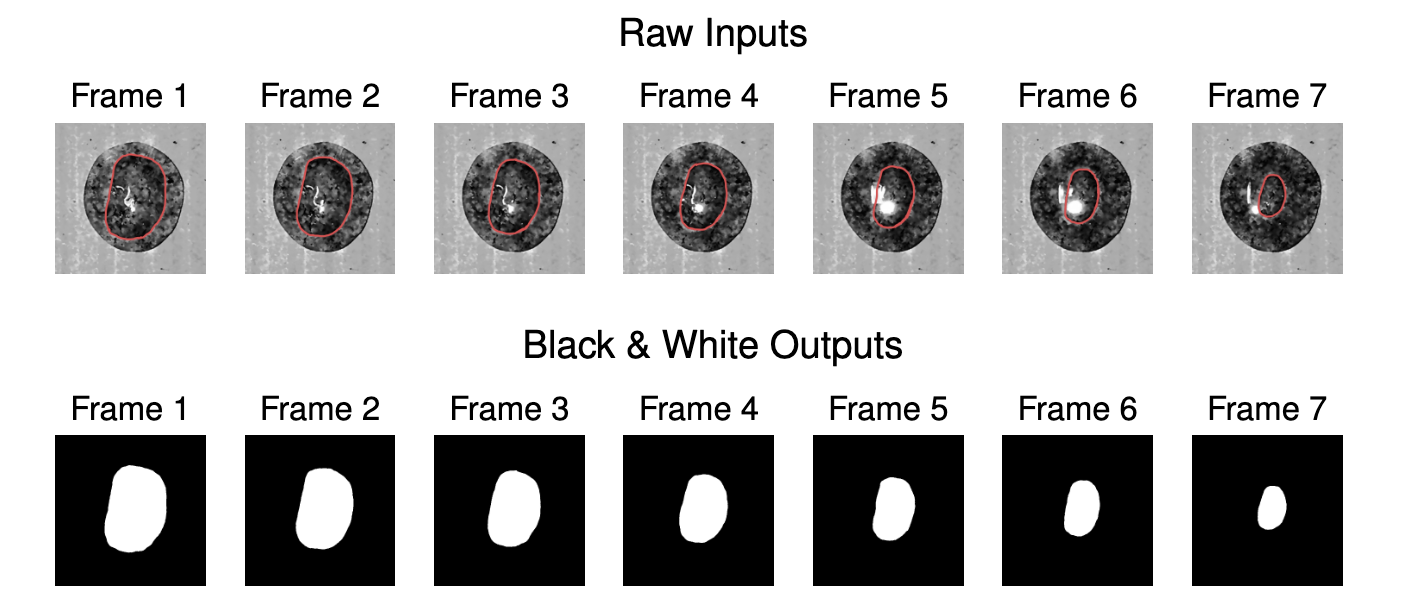}
    \caption{Example of one ice melting video. \textbf{Top: } Raw inputs. \textbf{Bottom: }Processed data.}
    \label{fig:ice}
\end{figure}

\noindent \textbf{Ice-melting dataset.}
The data for ice-melting dynamics were collected via experiments conducted in a controlled lab environment. Ice cubes were placed on a piece of kraft paper on a flat surface. Their melting was time-lapse video recorded from above using a camera on a tripod to capture a straight-on overhead view of the process. Frames were captured at two-minute intervals, and the boundary of the solid phase ice was manually marked with red outlines, as seen in the top panel of Figure \ref{fig:ice}. To obtain spatially aligned data, we selected only the videos in which the center of the ice remained fixed without rotation.

The annotated figures were then processed in Python, aiming to convert the ice and non-ice areas (which includes both the paper background and the water) into seven binary frames. The red outline was used to create binary images in which white and black correspond to ice and non-ice regions, respectively, shown in the bottom panel of Figure \ref{fig:ice}. The resulting binary frames, depicting the ice-shrinking process, were used as input data for the MBO networks.

\section{Results and Discussion}
In this section we present the results for synthetic and real data. We analyze the robustness of our models to noise, inaccurate kernel size estimation and dependence on the training dataset size.
\subsection{MBO network results on synthetic data}\label{MBOnetRes}
We present results for two dynamics: one generated by a standard Gaussian kernel with a threshold value of 0.2, and another by an MNIST kernel with a threshold of 0.5. These experiments are conducted under three conditions: no noise, Gaussian blur, and salt-and-pepper (SP) noise.  Recall that this method requires training a separate MBO network for each combination of kernel, threshold, and noise condition. Therefore, for brevity, we only show here these two examples, although similar results are observed for the other kernels and thresholds described in Section \ref{synt_data}. The MNIST kernels are particularly challenging due to their complex and diverse shapes, which make them  harder to recover.

For each scenario, we use 100 videos for training and 10 videos for testing. We trained the networks for 500 epochs using a smoothed thresholding function (sigmoid) with a steepness parameter of 100 as in equation \eqref{steepness} to avoid exploding gradients. We use the Heaviside function for testing.

All 100 training videos share the same kernel, threshold, and noise condition.
During training, the network is provided with the first frame and trained to accurately reconstruct the subsequent three frames (frames 2, 3, and 4). During testing, the first frame of an unseen video is given as input, and the MBO network is tasked with predicting the next six frames. The test error is computed based on the predicted six frames, assessing both the network’s ability to generalize to new videos and to extrapolate the dynamics over future time steps (as the network is trained on three frames, but tested on six). In the case of noisy conditions, both the input and target frames during training are noisy. At testing, the network is provided with a noisy first frame, and a Heaviside thresholding function is used to predict the next six frames. These predicted frames are then compared with the true, non-noisy frames to assess performance.

\begin{figure}[h!]
    \centering
    \includegraphics[width=0.8\linewidth]{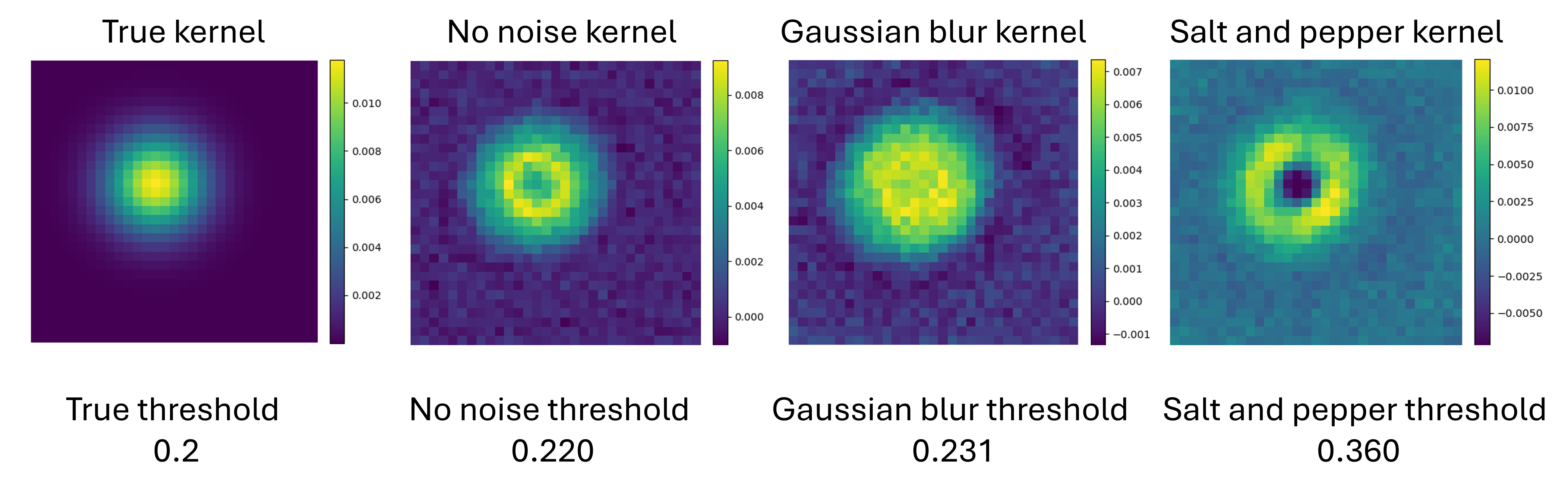}
    \caption{Reconstructed and true standard Gaussian kernel and threshold when using the MBO network. From left to right: the true kernel and threshold, the approximated kernel and threshold for the cases of no noise, Gaussian blur and salt-and-pepper noise.}
    \label{fig:SGkernel_recon}
\end{figure}

Figure \ref{fig:SGkernel_recon} illustrates the true and reconstructed kernels and thresholds for the standard Gaussian kernel with a threshold of 0.2 under three different conditions: no noise, Gaussian blur, and salt-and-pepper noise. In all cases, the location of the kernel's support is correctly identified; however, the support is slightly overestimated. This behavior is particularly noticeable in the Gaussian blur case, where the blurring effect tends to widen the perceived area of support. This overestimation is expected because Gaussian blur introduces smooth transitions, causing the model to interpret a wider range of values as part of the subject. As anticipated, the worst kernel reconstruction is observed in the salt-and-pepper noise case. This is due to the highly localized, extreme noise inherent in salt-and-pepper conditions, which introduces sharp, random disturbances. These disruptions make it difficult for the network to capture the precise kernel structure, leading to a more corrupted reconstruction. The threshold is accurately recovered in both the no-noise and Gaussian blur cases but is overestimated under salt-and-pepper noise. This overestimation likely occurs because the random pixel corruption forces the network to increase the threshold to account for the irregularities introduced by the noise.

\begin{figure}[h!]
    \centering
    \includegraphics[width=0.8\linewidth]{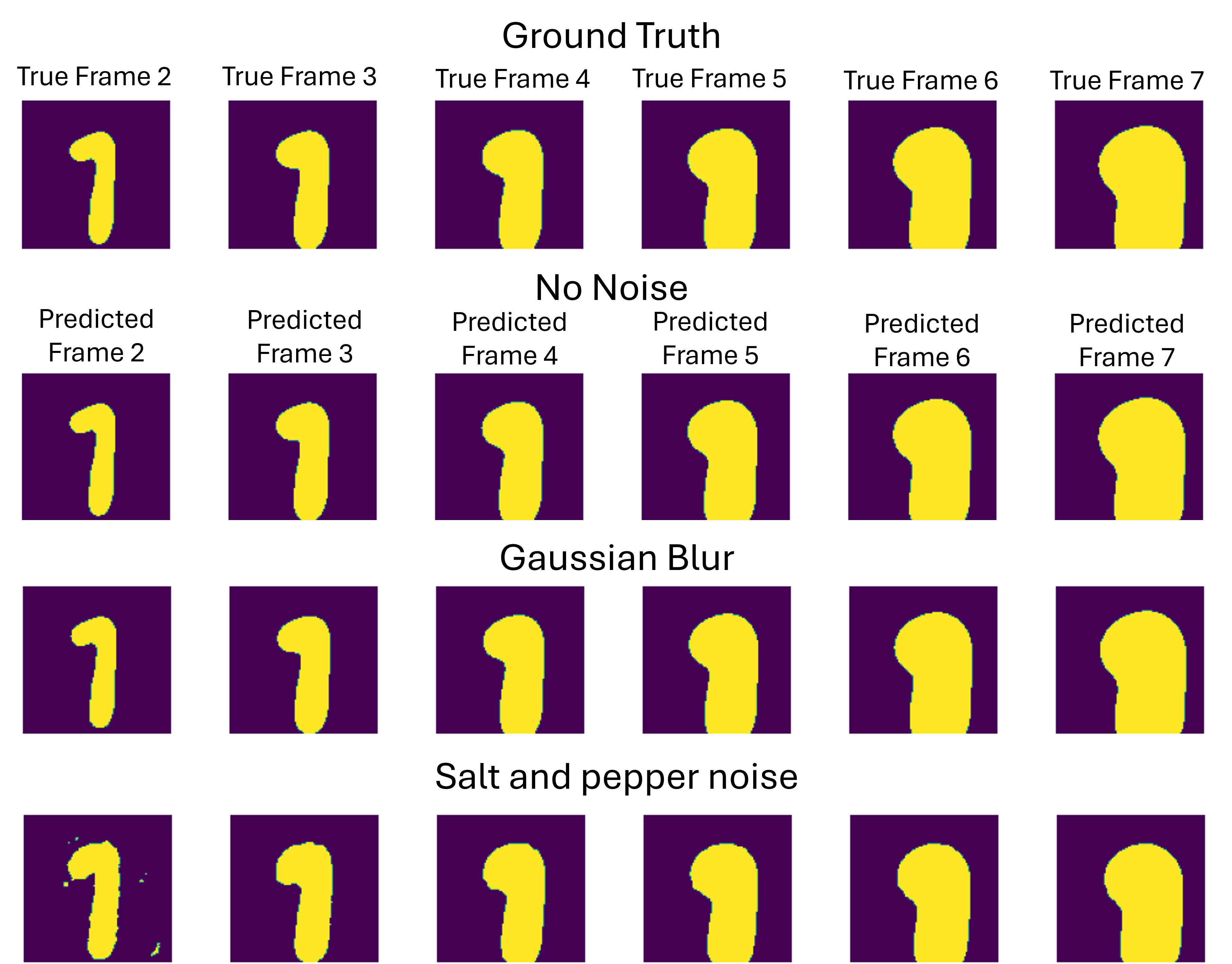}
    \caption{Reconstruction obtained using the MBO network on test data generated using standard Gaussian kernel and threshold 0.2. From left to right we show frames 1 through 7. From top to bottom we show the ground truth video and the reconstructed videos respectively in the case of no noise, Gaussian blur and salt-and-pepper noise.}
    \label{fig:SG_recon}
\end{figure}

In Figure \ref{fig:SG_recon}, we compare the ground truth and reconstructed videos across the three noise conditions. In both the no-noise and Gaussian blur scenarios, the video reconstructions are nearly perfect, with only minimal visible differences. However, some corruption is evident in the salt-and-pepper noise case, as expected. In the Gaussian blur case, while some smooth distortion is introduced, much of the underlying structure and dynamics of the subject is still retained. In contrast, salt-and-pepper noise creates high-frequency distortions that are much harder for the model to handle as they introduces artificial non-zero pixels that may be far from the original subject support (this is in contrast with the Gaussian blur case in which artificial non-zero pixels stay close the the original front). Nevertheless, in all cases, the overall dynamics is correctly reconstructed, particularly the location of the digit in the frame is always correctly determined even in extrapolated frames (frames 5,6,7).

\begin{figure}[h!]
    \centering
    \includegraphics[width=0.8\linewidth]{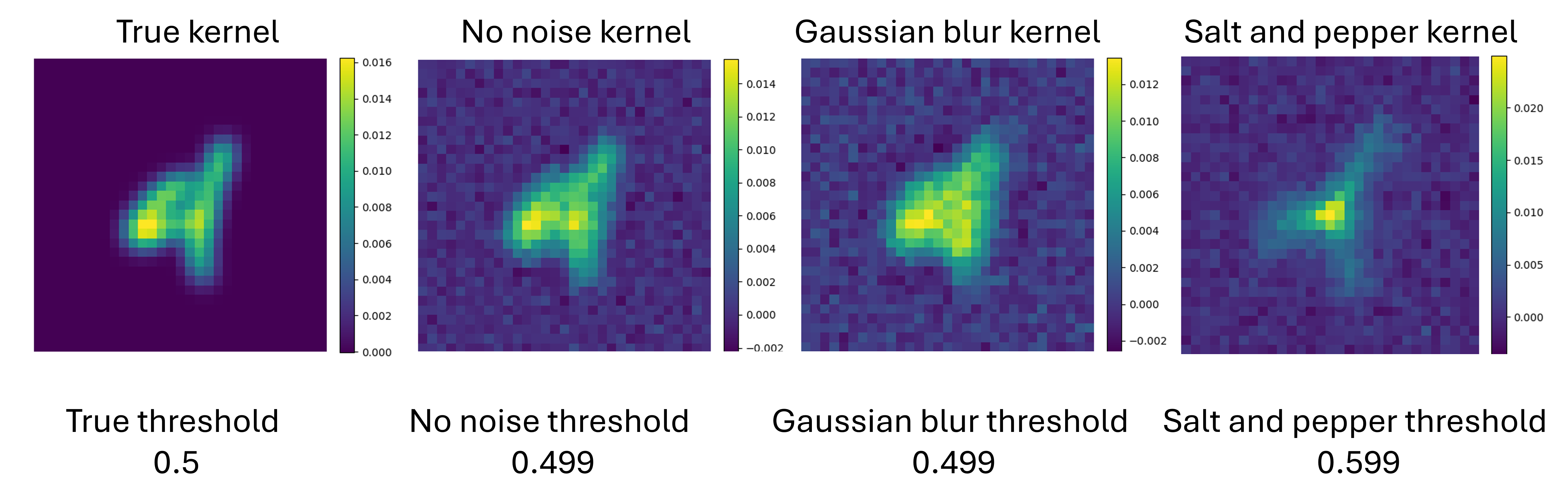}
    \caption{Reconstructed and true MNIST kernel and threshold when using MBO network. From left to right: the true kernel and threshold, the approximated kernel and threshold for the cases of no noise, Gaussian blur and salt-and-pepper noise.}
    \label{fig:MNISTkernel_recon}
\end{figure}

\begin{figure}[h!]
    \centering
    \includegraphics[width=0.8\linewidth]{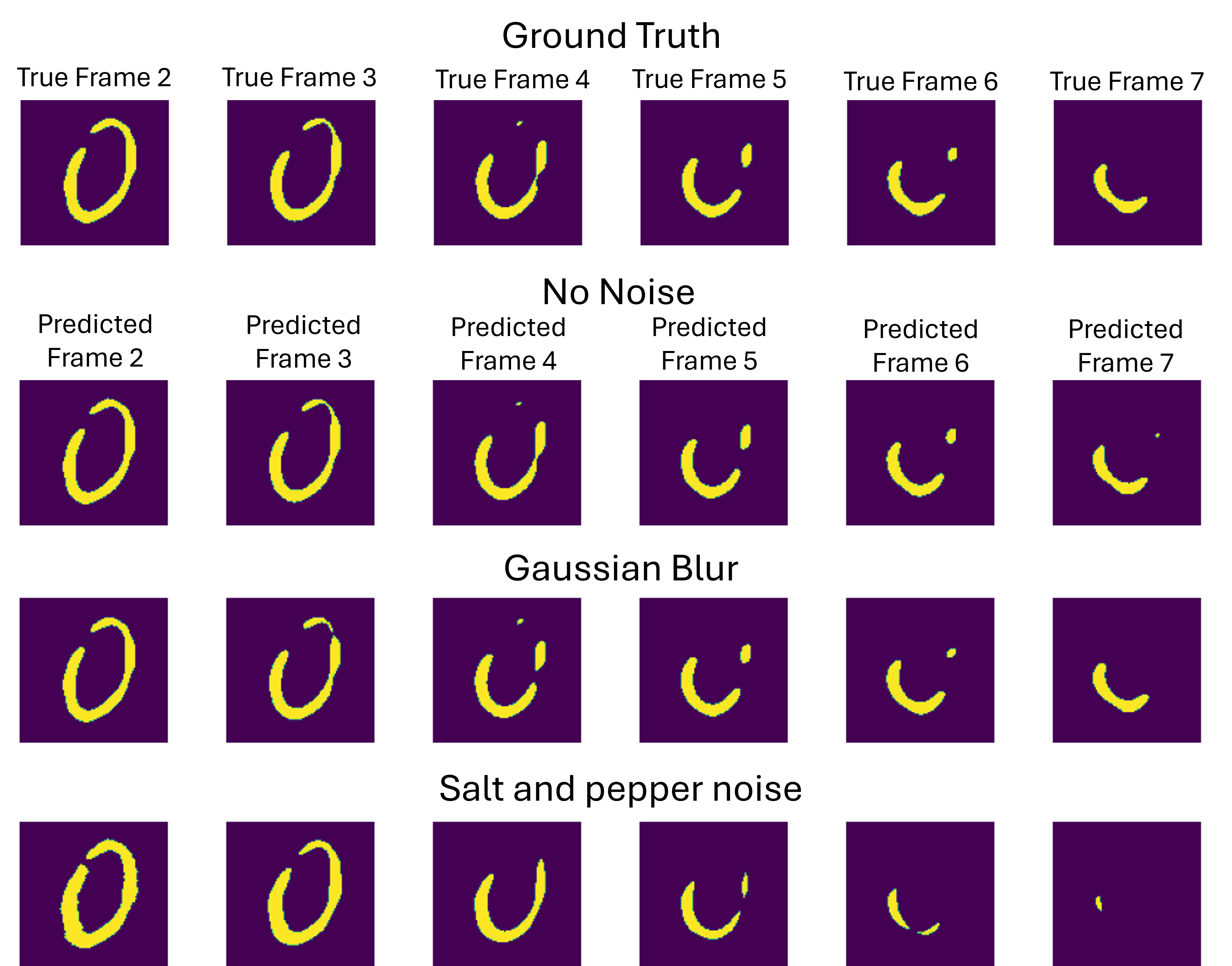}
    \caption{Reconstruction obtained using the MBO network on test data generated using MNIST kernel and threshold 0.5. From left to right we show frames 1 through 7. From top to bottom we show the ground truth video and the reconstructed videos respectively in the case of no noise, Gaussian blur and salt-and-pepper noise.}
    \label{fig:MNIST_recon}
\end{figure}

Finally, in Table \ref{GS_metrics} we report the relative MSE, the SSIM value and the Jaccard index on the test data. The test data is made of 10 previously unseen 7-frames videos. The results show that the no-noise condition achieves the best reconstruction, with a very low Relative MSE (0.293\%) and high SSIM (0.991) and Jaccard Index (99.707\%). Gaussian blur slightly degrades performance, reflected by a higher relative MSE (0.685\%) and a small decrease in SSIM (0.979) and Jaccard Index (99.318\%). The salt-and-pepper noise condition significantly worsens the performance, with a much larger relative MSE (9.557\%) and a noticeable drop in SSIM (0.743) and Jaccard Index (90.719\%). These results are consistent with the previous figures of reconstructed kernel, thresholds and videos. It is important to note that since Relative MSE calculates pointwise differences between images, it tends to be large for binary images. If a pixel that should be 1 is instead 0, or vice versa, it has a significant impact on the result, especially when the frame contains only a few non-zero pixels.
\begin{table}[H]
\centering
\caption{Performance metrics for MBO network on test data generated using a Gaussian kernel with threshold 0.2 under different noise conditions.}\label{GS_metrics}
\begin{tabular}{|c|c|c|c|}
\hline
\textbf{Noise Type} & \textbf{Relative MSE $\downarrow$} & \textbf{SSIM Value $\uparrow$} & \textbf{Jaccard Index $\uparrow$}  \\ \hline
No Noise           & 0.293\%                      & 0.991               & 99.707\%               \\ \hline
Gaussian Blur      & 0.685\%                      & 0.979               & 99.318\%               \\ \hline
Salt-and-Pepper Noise & 9.557\%                   & 0.743               & 90.719\%               \\ \hline
\end{tabular}
\end{table}
\newpage
In the case of the MNIST digit kernel with a threshold of 0.5, shown in Figure \ref{fig:MNISTkernel_recon}, the location of the kernel's support is correctly identified across all noise conditions. However, the kernel is accurately reconstructed only in the no-noise and Gaussian blur cases. Similar to previous results, the threshold is accurately recovered in these two conditions but is overestimated in the salt-and-pepper noise scenario.
In Figure \ref{fig:MNIST_recon}, we also display the reconstruction of a test video across all noise settings. Both the no-noise and Gaussian blur cases show near-perfect reconstruction, while artifacts appear in the salt-and-pepper noise case. Notably, in this case the last two predicted frames are almost empty, even though the ground truth still contains part of the digit, indicating the difficulty of handling salt-and-pepper noise.

In Table \ref{MNIST_metrics} we report the relative MSE, SSIM value, and Jaccard index on the test data. The results show that the no-noise condition yields the best performance, with a low relative MSE (1.669\%) and high SSIM (0.989) and Jaccard Index (98.351\%). The Gaussian blur condition results in slightly worse performance, with a higher relative MSE (3.295\%) and slight reductions in SSIM (0.980) and Jaccard Index (96.790\%). In the salt-and-pepper noise case, the performance declines significantly, with a large relative MSE (31.966\%) and noticeable drops in SSIM (0.823) and Jaccard Index (70.811\%). This decrease in accuracy is due to the fact that the last few predicted frames in the salt-and-pepper noise case often underestimate the amount of non-zero pixels present, as evident in Figure \ref{fig:MNIST_recon}.

These results reflect the increased complexity of the MNIST kernel compared to the simpler Gaussian kernel, as the MNIST digit has a more intricate support structure and a varied pixel intensity distribution. Additionally, since the subject in these videos shrinks over time because of the 0.5 threshold, many frames contain very few non-zero pixels, which as explained in the previous example, tend to increase the error. In particular, the relatve MSE tends to be higher in such cases due to the significant impact of pointwise differences between 1s and 0s, especially when the true frames have sparse non-zero pixels.

\begin{table}[H]
\centering
\caption{Performance metrics for MBO network on test data generated using an MNIST kernel with threshold 0.5 under different noise conditions.}
\label{MNIST_metrics}
\begin{tabular}{|c|c|c|c|}
\hline
\textbf{Noise Type} & \textbf{Relative MSE $\downarrow$} & \textbf{SSIM Value $\uparrow$} & \textbf{Jaccard Index $\uparrow$}  \\ \hline
No Noise           & 1.669\%                      & 0.989               & 98.351\%               \\ \hline
Gaussian Blur      & 3.295\%                      & 0.980               & 96.790\%               \\ \hline
Salt-and-Pepper Noise & 31.966\%                   & 0.823               & 70.811\%               \\ \hline
\end{tabular}
\end{table}
\newpage
\subsubsection{Robustness to Inaccurate Kernel Dimension}\label{MBO_kerdim}
In practice, the true kernel and thresholds used to generate a threshold dynamic video are not known. As a result, during training, we must estimate the kernel size, which often differs from the actual size underlying the ground truth dynamics. In this section, we examine the robustness of our method to variations in the assumed kernel size. Specifically, we generate data using a standard 31x31 Gaussian kernel. However, during training and testing, we experiment with kernels of different sizes: 15x15 (underestimated kernel size) and 51x51 (overestimated kernel size). For brevity, we present results in the case of the standard Gaussian kernel used in Section \ref{MBOnetRes} with threshold 0.2. In the following we show that the model is robust to inaccuracies in the assumed kernel size and that overestimating the kernel size yields results comparable in accuracy to those obtained with the exact kernel size.

\begin{figure}[h!]
    \centering
    \includegraphics[width=0.8\linewidth]{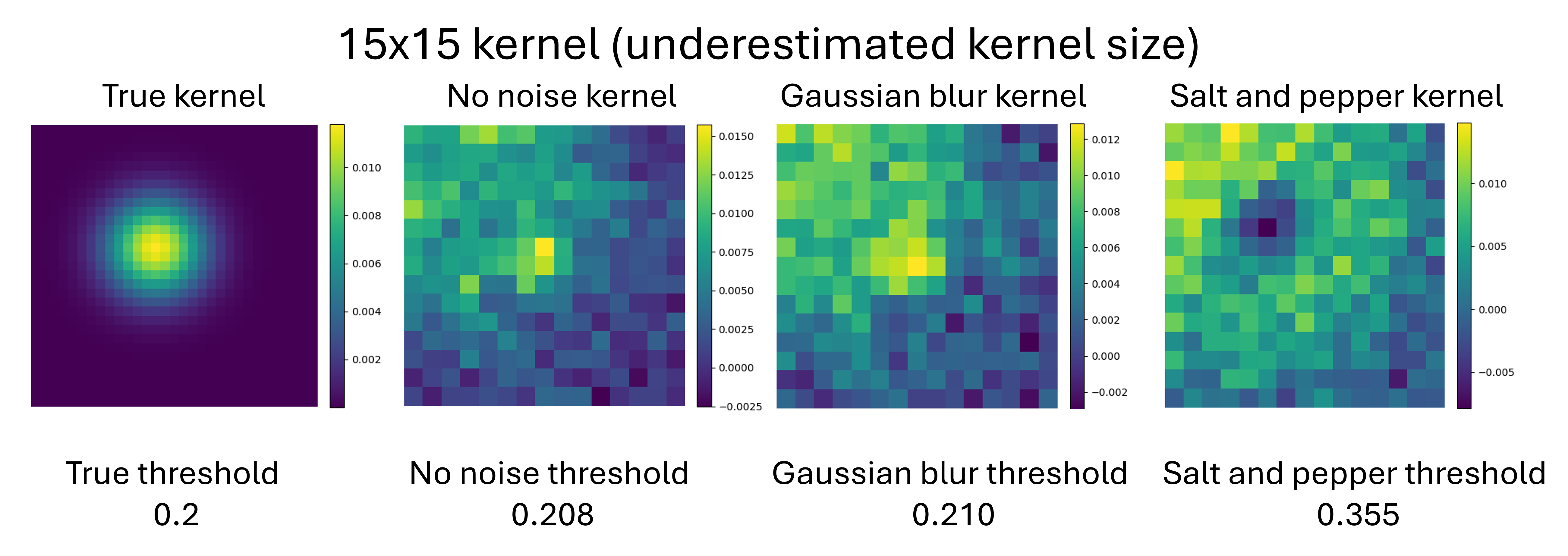}
    \caption{Reconstructed and true kernel and threshold using MBO network when the kernel size is underestimated during training and testing to be 15x15 instead of the true size 31x31. From left to right: the true 31X31 kernel and
threshold, the approximated 15x15 kernels and thresholds for the cases of no noise, Gaussian blur and salt-and-pepper noise.}
    \label{fig:SG15_ker}
\end{figure}

When using an underestimated 15x15 kernel, we observe from Figure \ref{fig:SG15_ker} that the reconstructed kernel resembles a zoomed-in version of the original 31x31 kernel, particularly around its central 15x15 block. This result is intuitive because the video frames are produced by convolution, and when learning a smaller kernel, the network tends to focus on capturing the central part of the kernel. This occurs because the central region of the kernel often contains the most significant information for convolution, influencing the dynamics of the video more strongly than the outer regions.

The threshold approximations remain fairly accurate, but the approximations of the kernels are not as precise as in the case where the exact kernel size was used. Consequently, the performance metrics deteriorate slightly, as reflected in the results shown in Table \ref{SG15_metrics}. In the no-noise condition, the relative MSE is higher (0.606\%) compared to the exact kernel size case in Section \ref{MBOnetRes}, and both the SSIM (0.982) and Jaccard Index (99.396\%) are marginally lower. Similarly, in the Gaussian blur condition, the relative MSE increases (0.796\%), while the SSIM (0.976) and Jaccard Index (99.207\%) also decline slightly. As in previous cases, the salt-and-pepper noise condition leads to the largest drop in performance, with a significantly higher relative MSE (11.461\%) and noticeably lower SSIM (0.708) and Jaccard Index (88.896\%). These results indicate that while the network can still capture the core dynamics of the kernel and thresholds, using a smaller estimated kernel introduces some inaccuracies, particularly in noisier conditions.

\begin{table}[h!]
\centering
\caption{Performance metrics on test data for MBO network when using an underestimated 15x15 kernel size under different noise conditions.}
\label{SG15_metrics}
\begin{tabular}{|c|c|c|c|}
\hline
\textbf{Noise Type} & \textbf{Relative MSE $\downarrow$} & \textbf{SSIM Value $\uparrow$} & \textbf{Jaccard Index $\uparrow$}  \\ \hline
No Noise           & 0.606\%                      & 0.982               & 99.396\%               \\ \hline
Gaussian Blur      & 0.796\%                      & 0.976               & 99.207\%               \\ \hline
Salt-and-Pepper Noise & 11.461\%                   & 0.708               & 88.896\%               \\ \hline
\end{tabular}
\end{table}

\begin{figure}[h!]
    \centering
    \includegraphics[width=0.8\linewidth]{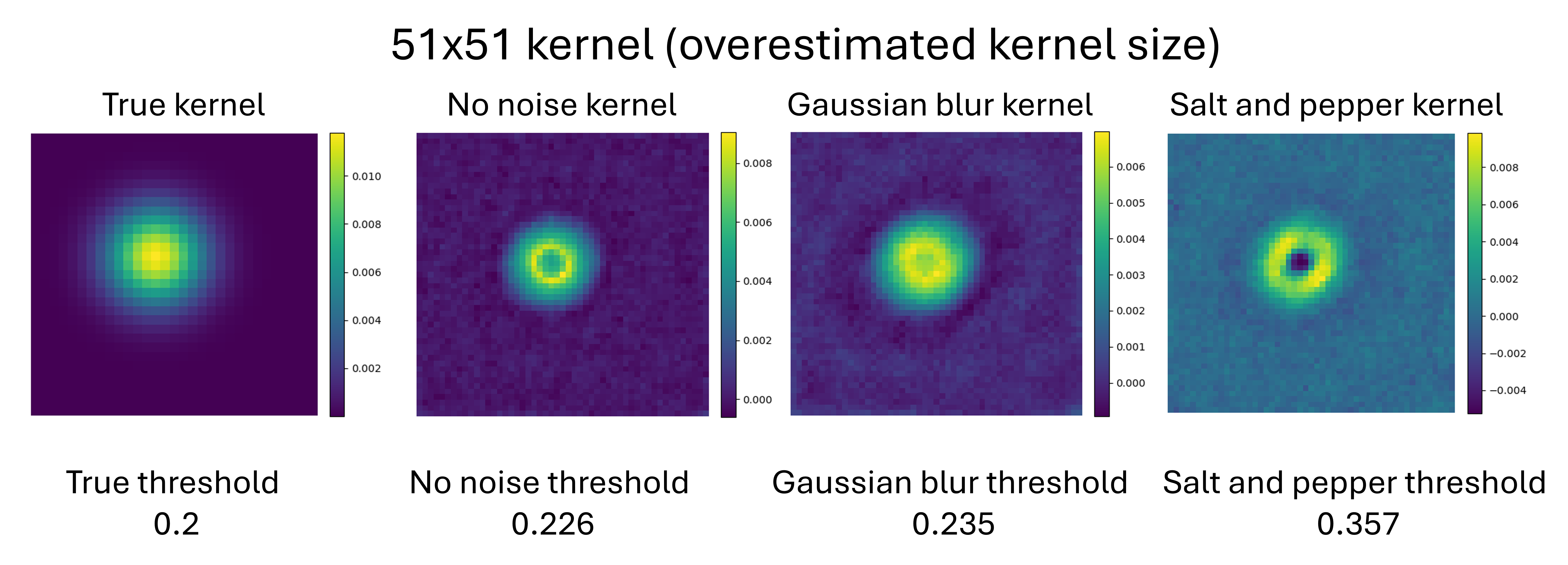}
    \caption{Reconstructed and true kernel and threshold using MBO network when the kernel size is overestimated to be 51x51 during training and testing instead of the true size 31x31. From left to right: the true 31X31 kernel and
threshold, the approximated 51x51 kernels and thresholds for the cases of no noise, Gaussian blur and salt-and-pepper noise.}
    \label{fig:SG51_ker}
\end{figure}

When the kernel size is overestimated to be 51x51, the reconstructed kernels appear as zoomed-out versions of the original 31x31 kernel as seen in Figure \ref{fig:SG51_ker}. The kernel reconstructions capture the full support of the original kernel, and the pixel intensity distribution across the kernel is visually accurate. The location of the kernel’s support is consistently recovered in all conditions, similar to the case of the smaller kernel. Notably, the thresholds are also accurately approximated, matching the performance observed with the exact 31x31 kernel case.

The metrics in Table \ref{SG51_metrics} also reflect that this overestimation of the kernel size does not significantly degrade performance. In the no-noise condition, the relative MSE remains low (0.394\%), with high SSIM (0.987) and Jaccard Index (99.607\%), indicating good reconstruction quality. For the Gaussian blur case, there is a slight increase in relative MSE (0.978\%) and a minor reduction in SSIM (0.969) and Jaccard Index (99.027\%), but the overall performance remains strong. The salt-and-pepper noise condition, as usual, results in the largest performance drop, but still comparable with the 31x31 kernel case, with a relative MSE of 9.834\%, SSIM of 0.738, and Jaccard Index of 90.479\%. In all noise conditions the results when using a 51x51 kernel are as accurate, or slightly less accurate, than the 31x31 kernel case. This is expected because overestimating the kernel size allows for the network to recover a kernel that is structurally more similar to the original 31x31 one, in contrast with the 15x15 case.

Is it worth noting that overestimating the kernel size may require additional training time due to the increased number of parameters. However, this approach can be advantageous when the true kernel dimension is unknown, as it allows the network to fully capture the kernel’s support.\\

\begin{table}[H]
\centering
\caption{Performance metrics on test data for MBO network when using an overestimated 51x51 kernel size under different noise conditions.}
\label{SG51_metrics}
\begin{tabular}{|c|c|c|c|}
\hline
\textbf{Noise Type} & \textbf{Relative MSE $\downarrow$} & \textbf{SSIM Value $\uparrow$} & \textbf{Jaccard Index $\uparrow$}  \\ \hline
No Noise           & 0.394\%                      & 0.987               & 99.607\%               \\ \hline
Gaussian Blur      & 0.978\%                      & 0.969               & 99.027\%               \\ \hline
Salt-and-Pepper Noise & 9.834\%                    & 0.738               & 90.479\%               \\ \hline
\end{tabular}
\end{table}

\subsubsection{Learning with one short video}\label{sec:one_vid}
In practical scenarios,  access to multiple videos generated by the same kernel and threshold may not be feasible. However, since the primary goal of the network is to learn just two elements---a threshold value and a 31x31 kernel---it may not require a large number of training videos (e.g., the 100 videos used for training in the previous sections). In this section, we explore this intuition by investigating the network’s performance when trained using only \textbf{one video} with 4 frames. With limited training data, we observed that for the loss to plateau a larger number of epochs was needed. In this experiment we train for 2000 epochs instead of 500. Notably, since only one video is used for training, the overall training time is shorter than in the previous example.

During testing, the network is given the first frame from 10 different videos (generated using the same kernel and threshold as the training video ones) and tasked with predicting the subsequent 6 frames. The test error is then evaluated to assess the effectiveness of training with limited data. Again for brevity, we present results only for the Gaussian kernel with a threshold of 0.2, although similar results can be obtained with other kernels and thresholds.

Compared to the results for the standard Gaussian kernel with threshold 0.2, where 100 training videos were used, we observe (see Table \ref{tab:one_video_results}) an expected increase in errors across all noise conditions when only one video is used for training. The relative MSE is significantly larger, especially in the salt-and-pepper noise case, and both SSIM and Jaccard Index values show a noticeable drop. This discrepancy is due to the reduced amount of data---training with only one video provides less information for the network to generalize effectively. Nonetheless, the network still performs well, particularly in the no-noise and Gaussian blur conditions. The error remains quite low, which is impressive given that the network is predicting unseen frames and extrapolating up to frame 7.

\begin{figure}[h!]
    \centering
    \includegraphics[width=0.8\linewidth]{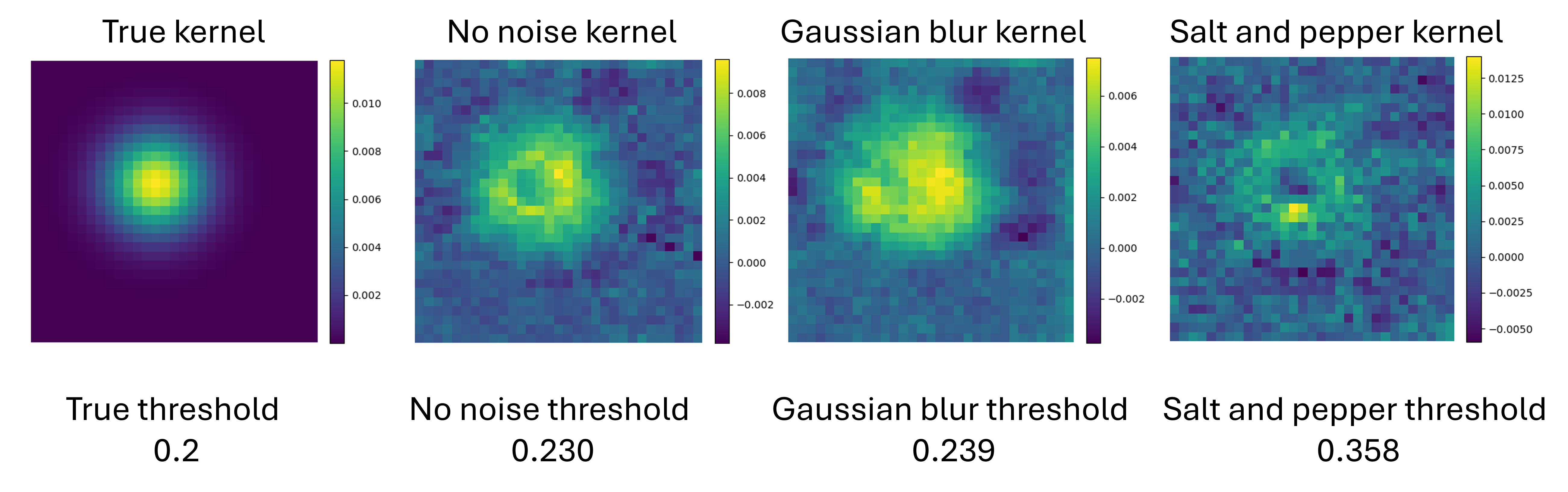}
    \caption{Reconstructed and true standard Gaussian kernel and threshold using MBO network when only one video is used for training. From left to right the true kernel and threshold, the approximated kernel and threshold for the cases of no noise, Gaussian blur and salt-and-pepper noise. Training was done using only one 4 frames video. }
    \label{fig:1vidSGkernel_recon}
\end{figure}

The impact of using only one training video is also evident in the kernel and threshold reconstructions in Figure \ref{fig:1vidSGkernel_recon}. The support of the reconstructed kernel is more spread out compared to the case with a large training dataset, and in the salt-and-pepper noise condition, the pixel intensity distribution is visibly inaccurate. For an example of the test video reconstruction corresponding to these kernels we refer to Appendix~\ref{one_short_app}.

\begin{table}[H]
\centering
\caption{Performance metrics on test data for MBO network when only one video is used for training.}
\label{tab:one_video_results}
\begin{tabular}{|c|c|c|c|}
\hline
\textbf{Noise Type} & \textbf{Relative MSE $\downarrow$} & \textbf{SSIM Value $\uparrow$} & \textbf{Jaccard Index $\uparrow$}  \\ \hline
\text{No Noise} & 1.560\% & 0.958 & 98.454\% \\ \hline
\text{Gaussian Blur} & 2.842\% & 0.925 & 97.198\% \\ \hline
\text{Salt-and-Pepper Noise} & 12.474\% & 0.673 & 88.071\% \\ \hline
\end{tabular}
\end{table}

\subsection{Meta-learning MBO results on synthetic data}\label{metaL}
In this section, we present the results for the meta-learning MBO network. The data used is generated by choosing the threshold values among $0.2$, $0.3$, $0.5$, and $0.6$. For each fixed threshold, a kernel is sampled from the five classes introduced in Section \ref{synt_data}. This process is repeated five times, resulting in 100 distinct combinations of kernels and thresholds. For each combination, 30 videos are generated, leading to a total of 3000 videos. The 3000 videos are split into 90\% for training and 10\% for testing. In each noise condition (no noise, Gaussian blur, and salt-and-pepper noise), we train the network using only the first four frames of each video, while for testing, we generate predictions up to frame 7. Training is performed for 500 epochs with a smoothed thresholding function (sigmoid) with a steepness parameter of 100 as in equation \eqref{steepness} to avoid exploding gradients.

Table \ref{tab:mean_metrics_meta} presents the performance metrics on the test data for each noise condition. Note that the test videos were never seen during training, and while training is done using four frames, seven frames are predicted for testing.
As seen from the table, the no-noise condition achieves the best performance, with a low relative MSE (4.31\%), high SSIM (0.961), and a strong Jaccard Index (94.1\%). The Gaussian blur condition leads to slightly degraded performance, with a higher relative MSE (8.53\%) and a decrease in both SSIM (0.915) and Jaccard Index (89.6\%). The salt-and-pepper noise condition shows the largest drop in performance, reflected in the significantly higher relative MSE (37.61\%) and lower SSIM (0.774) and Jaccard Index (58.2\%). These results align with our expectations, given the increasing difficulty in reconstructing the test frames under the various noise conditions. 

\begin{table}[H]
\centering
\caption{Performance metrics on test data for meta-learning MBO network for different noise conditions.}
\label{tab:mean_metrics_meta}
\begin{tabular}{|c|c|c|c|}
\hline
\textbf{Noise Type} & \textbf{Relative MSE $\downarrow$} & \textbf{SSIM Value $\uparrow$} & \textbf{Jaccard Index $\uparrow$} \\ \hline
\text{No Noise} & 4.31\% & 0.961 & 94.1\% \\ \hline
\text{Gaussian Blur} & 8.53\% & 0.915 & 89.6\%  \\ \hline
\text{Salt-and-pepper Noise} & 36.0\% & 0.779 & 59.9\%  \\ \hline
\end{tabular}
\end{table}

We also explicitly note the main difference between this approach and the previously presented one. In the MBO network from Section \ref{MBOnetRes}, the results appear to be more accurate compared to those obtained here. However, in that previous case, the network learned only one combination of kernel and threshold. In contrast, the current approach involves training a single architecture on videos generated from 100 different combinations of kernels and thresholds. If we had used the same MBO network as before, we would have required 100 independent networks to obtain the same results shown here. 

\begin{figure}[h!]
    \centering
    \includegraphics[width=0.8\linewidth]{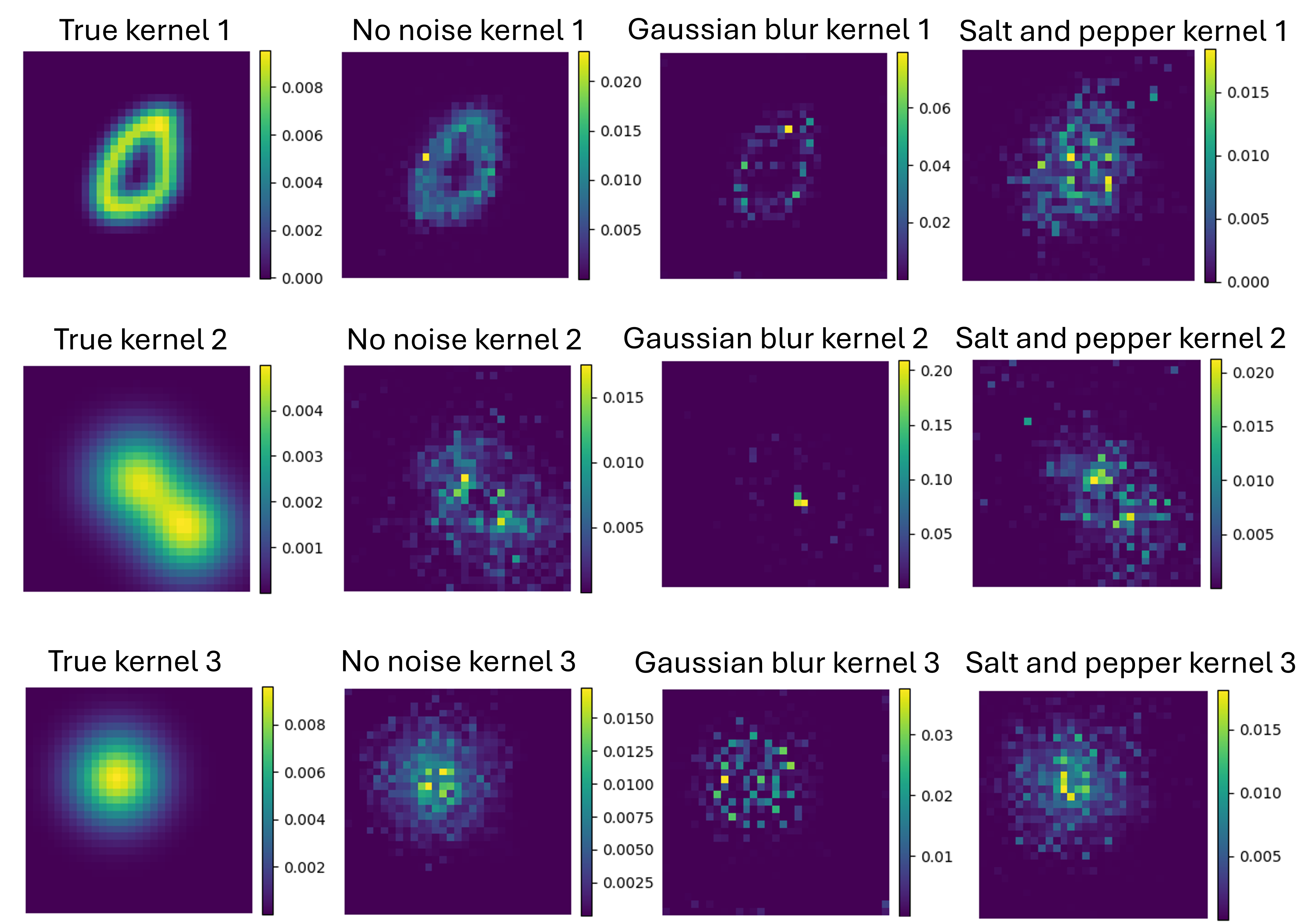}
    \caption{Reconstructed and true kernels when using meta-learning MBO network on three test videos. From left to right: the true kernels, the approximated kernels for the cases of no noise, Gaussian blur and salt-and-pepper noise.}
    \label{fig:kernel_meta}
\end{figure}

Figure \ref{fig:kernel_meta} and Table \ref{thresholds_meta} show respectively true and reconstructed kernels and thresholds for three randomly picked test videos. In all cases we can see that the kernel location is correctly identified, while the pixel intensity of the kernel is often underestimated, especially in presence of noise. Similarly, the threshold reconstruction is less accurate when noise is present in the data. Notably, the fact that the meta-leraning MBO network has access to multiple videos generated from a variety of kernels results in more accurate kernel reconstruction at testing time compared with the results obtained by the MBO network when trained and tested on one video (see Section \ref{sec:one_vid}). Comparing Figure~\ref{fig:kernel_meta} and Figure~\ref{fig:1vidSGkernel_recon} we can see that in both cases the support of the kernel is accurately recovered, but the intensity (especially of the background) is much more accurate when using the meta-learning MBO.

\begin{figure}[h!]
    \centering
    \includegraphics[width=0.8\linewidth]{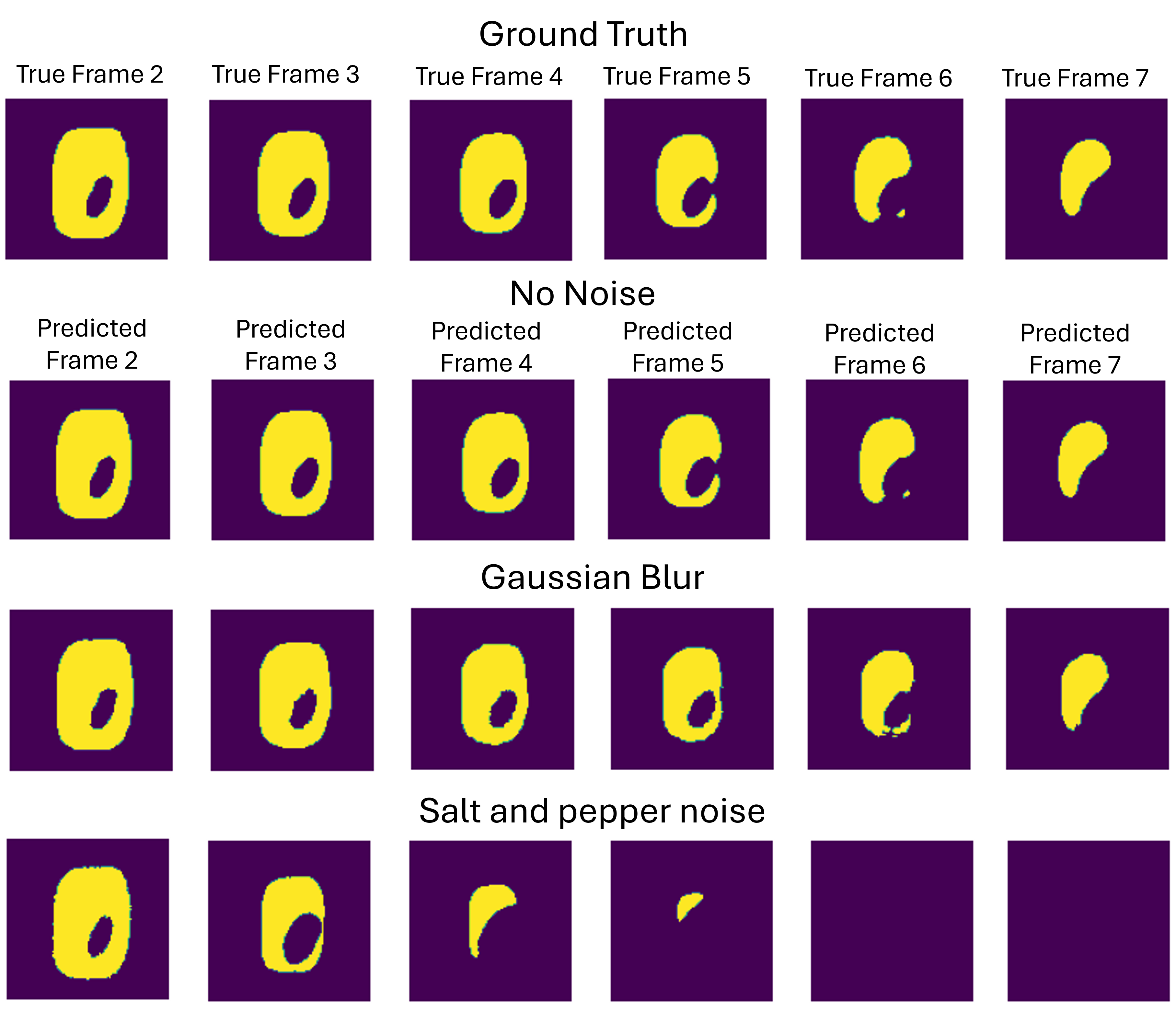}
    \caption{Reconstructed and true test video using meta-learning MBO. These dynamics were generated the kernel 1 (MNIST 0 digit) from the previous figure and threshold 0.6. From left to right we show frames 1 through 7. From top to bottom we show the ground truth video and the reconstructed video respectively in the case of no noise, Gaussian blur and salt-and-pepper noise. This is a test video never seen during training.}
    \label{fig:meta0.6_recon}
\end{figure}

In Figure \ref{fig:meta0.6_recon}, we show the reconstruction of a test video corresponding to the first kernel (MNIST 0 digit) and threshold 0.6 across noise conditions. Despite the inaccuracy in the pixel intensity of the reconstructed kernel, the resulting videos in the case of no noise and Gaussian blur are very similar to the ground truth. Conversely, in the salt-and-pepper noise condition, the inaccurately reconstructed threshold (0.8285 instead of 0.6) leads to a dynamic that shrinks faster than  it is supposed to, resulting in empty final frames. Again this confirms what observed in previous examples that salt-and-pepper noise is a challenging setting for this technique.

\begin{table}[h!]
\centering
\caption{True and learned thresholds for three test videos when using meta-learning MBO network in different noise conditions.}
\label{thresholds_meta}
\begin{tabular}{|c|c|c|c|}
\hline
\textbf{True Threshold} & \textbf{\begin{tabular}[c]{@{}c@{}}No noise\\ threshold\end{tabular}} & \textbf{\begin{tabular}[c]{@{}c@{}}Gaussian Blur\\ threshold\end{tabular}} & \textbf{\begin{tabular}[c]{@{}c@{}}Salt-and-Pepper noise\\ threshold\end{tabular}} \\ \hline
0.6                     & 0.6036                                                                & 0.5960                                                                     & 0.8285                                                                             \\ \hline
0.3                     & 0.3043                                                                & 0.1605                                                                     & 0.4813                                                                             \\ \hline
0.6                     & 0.6033                                                                & 0.5847                                                                     & 0.8220                                                                             \\ \hline
\end{tabular}
\end{table}

\newpage
\subsubsection{Robustness to Inaccurate Kernel Dimension}
Similar to the MBO network, to train the meta-learning MBO network, an a priori estimation of the kernel size is necessary. The true kernel used to generate the synthetic data is 31x31 pixels. In this section, we investigate the impact of using an underestimated kernel size of 15x15 and an overestimated kernel size of 51x51 during training. We analyze how these choices influence the reconstructed kernel and the accuracy of the generated dynamics. We show here results only for the cases of no noise and Gaussian Blur noise since in the case of salt-and-pepper noise we obtained large reconstruction errors even for the exact kernel size (see Table \ref{tab:mean_metrics_meta}).

When using the 15x15 kernel, the reconstructed kernel resembles the true kernel but appears as a zoomed-in version, similar to what we observed for the MBO network in Section \ref{MBO_kerdim}. For brevity, we omit the images of the kernels. This effect occurs because the smaller kernel cannot fully capture the true kernel’s support within the 15x15 domain. Consequently, parts of the kernel are lost, leading to an incomplete representation of the dynamics. As a result, the error increases in this case (see Table~\ref{tab15}) compared to the 31x31 kernel case (see Table \ref{tab:mean_metrics_meta}). This outcome aligns with the finding for the MBO network in Section~\ref{MBO_kerdim}, where underestimating the kernel size also produced higher errors due to the inability to capture the full extent of the kernel’s support. Table~\ref{tab15} shows the error in the reconstruction of test videos for the 15x15 kernel case. 

\begin{table}[h!]
\centering
\caption{Performance metric on test data of meta-learning MBO network when using a kernel of size 15x15 (underestimated).}
\label{tab15}
\begin{tabular}{|c|c|c|c|}
\hline
Noise Type    & Relative MSE $\downarrow$ & SSIM Value $\uparrow$ & Jaccard Index $\uparrow$ \\ \hline
No Noise   & 5.60\%                    & 0.958                 & 92.2\%                   \\ \hline
Gaussian Blur & 9.81\%                    & 0.908                 & 88.0\%                   \\ \hline
\end{tabular}

\end{table}

On the other hand, when using the 51x51 kernel, the reconstructed kernel resembles a zoomed-out version of the true kernel. The entire support of the true kernel is contained within the 51x51 domain. However, the additional pixels surrounding the kernel are close to zero, which creates a ``padding'' effect. This padding slightly dilutes the signal, causing some inaccuracies in the generated dynamics. Despite this, the error remains comparable to the correct 31x31 case because the full support of the kernel is preserved. Table \ref{tab51} shows the error in the reconstruction of test videos for the 51x51 kernel case. 
\begin{table}[h!]
\centering
\caption{Performance metric on test data of meta-learning MBO network when using a kernel of size 51x51 (overestimated).}
\label{tab51}
\begin{tabular}{|c|c|c|c|}
\hline
Noise Type    & Relative MSE $\downarrow$ & SSIM Value $\uparrow$ & Jaccard Index $\uparrow$ \\ \hline
No Noise      & 5.31\%                    & 0.952                 & 92.8\%                   \\ \hline
Gaussian Blur & 9.21\%                    & 0.909                 & 88.7\%                   \\ \hline
\end{tabular}

\end{table}

In summary, these results demonstrate that underestimating the kernel size (15x15) leads to a increase in error, as the truncated kernel fails to represent the full dynamics. In contrast, overestimating the kernel size (51x51) results in only a marginal increase in error, since the extra near-zero pixels around the kernel have less influence on the dynamics than losing part of the kernel’s support. These trends are consistent with the behavior observed in Section~\ref{MBO_kerdim}, showing that, if no information is known about the true kernel, a larger kernel size would result in better reconstruction.

\subsection{Learning from real data}
In this section, we present the results of our method on the real-world datasets introduced in Section \ref{real_data}. More specifically, we evaluate performance of our methods on two datasets: one containing 5 videos of the expansion of fire fronts and another containing 5 videos of melting ice. For the ice-melting videos, we expect the dynamics to be produced by a Gaussian kernel with a threshold of 0.5, while for the fire videos, we anticipate a kernel represented by the indicator function of a disc and a threshold smaller than 0.5. It is important to note that we expect higher accuracy on the ice-melting videos compared to the fire videos; this is because the ice-melting videos were recorded under controlled lab conditions (see Section \ref{real_data}), while the fire videos were sourced from real-world footage. Consequently, the fire dynamics may have been affected by external factors such as wind, weather, or physical barriers (e.g., lakes or mountains), none of which are taken into account by the model.

We compare the MBO network and the meta-learning MBO network results on real data in terms of \textit{generalization} and \textit{extrapolation}. For generalization we test the ability of the two networks to reconstruct videos never seen during training and possibly generated with unseen kernels. For extrapolation we test the ability of the networks to generate accurate future frames (i.e., we train with frames 1-4 and test on frames 5-7).

\subsubsection{Ice Dataset}
\textbf{Generalization.} For the MBO network, we trained the model using one ice-melting video to learn kernel and threshold. These learned parameters were then used, given the first frame, to predict the remaining frames for the other 4 test videos. The error was calculated by comparing the true and predicted frames of these test videos. This tests the ability of the MBO network to generalize to unseen videos since only one video was used for training. Note that since the MBO network is trained with only one ice-meting video, its success in generalization strongly depends on the assumption that all ice-melting videos were generated with the same kernel and threshold. While in theory this is the case, in practice kernels and threshold may vary slightly across videos because of noise, experimental setup etc. We expect this to strongly impact reconstruction for this case.

For the meta-learning MBO network, we directly used the pre-trained model from Section~\ref{metaL} without any fine-tuning on the real data. Since the pre-trained model was trained on videos corresponding to a variety of kernel and threshold combinations, including the Gaussian kernel with threshold 0.5, we expect it to be able to approximately reconstruct the dynamics of the ice-melting videos. We show test results when using the model pre-trained on noiseless data and on blurry data. Since in presence of salt-and-pepper noise the method did not provide good results on synthetic data we omit its application to real data. Note that the meta-learning MBO-network was never shown any ice-melting videos during training so it has to generalize to completely unseen dynamics.

To ensure a fair comparison, we evaluated both the meta-learning architectures on the same 4 test ice-melting videos used in testing for the MBO network. The results are shown in Table \ref{tab:ice_recon_metrics}.

\begin{table}[h!]
\centering
\caption{Performance comparison of the MBO network and the meta-learning MBO in the reconstruction of the 7-frames ice-melting videos.}
\label{tab:ice_recon_metrics}
\begin{tabular}{|c|c|c|c|}
\hline
\textbf{Model} & \textbf{Relative MSE $\downarrow$} & \textbf{SSIM Value $\uparrow$} & \textbf{Jaccard Index $\uparrow$} \\ \hline
\text{MBO Network} & 48.306\% & 0.859 & 52.440\% \\ \hline
\text{Meta-learning MBO (No Noise)} & 19.9\% & \textbf{0.936} & 79.6\% \\ \hline
\text{Meta-learning MBO (Gaussian Blur)} & \textbf{18.65\%} & 0.933 & \textbf{80.33\%} \\ \hline
\end{tabular}

\end{table}

For the ice-melting dataset, the results show that the meta-learning MBO network trained on videos with Gaussian blur outperforms in terms of generalization both the MBO network and the meta-learning MBO trained on noiseless data. Specifically, the blurred model achieves the lowest relative MSE of 18.65\% and the highest Jaccard Index of 80.33\%. This result is reasonable because real-world ice-melting videos likely contain inherent noise from environmental factors such as lighting, camera quality, or slight inconsistencies in the experimental setup. Consequently, the Meta-learning MBO trained with blurred (noisy) data is more robust to these real-world imperfections, enabling it to produce more accurate reconstructions. It is also expected that the MBO network would not perform as well in this scenario since it assumes that all videos share the same kernel and threshold, leading to a more rigid framework. In contrast, the meta-learning MBO's flexibility given by its training across multiple combinations of kernels and thresholds make it better suited for handling the complexities of real-world data.

\begin{figure}
    \centering
    \includegraphics[width=0.8\linewidth]{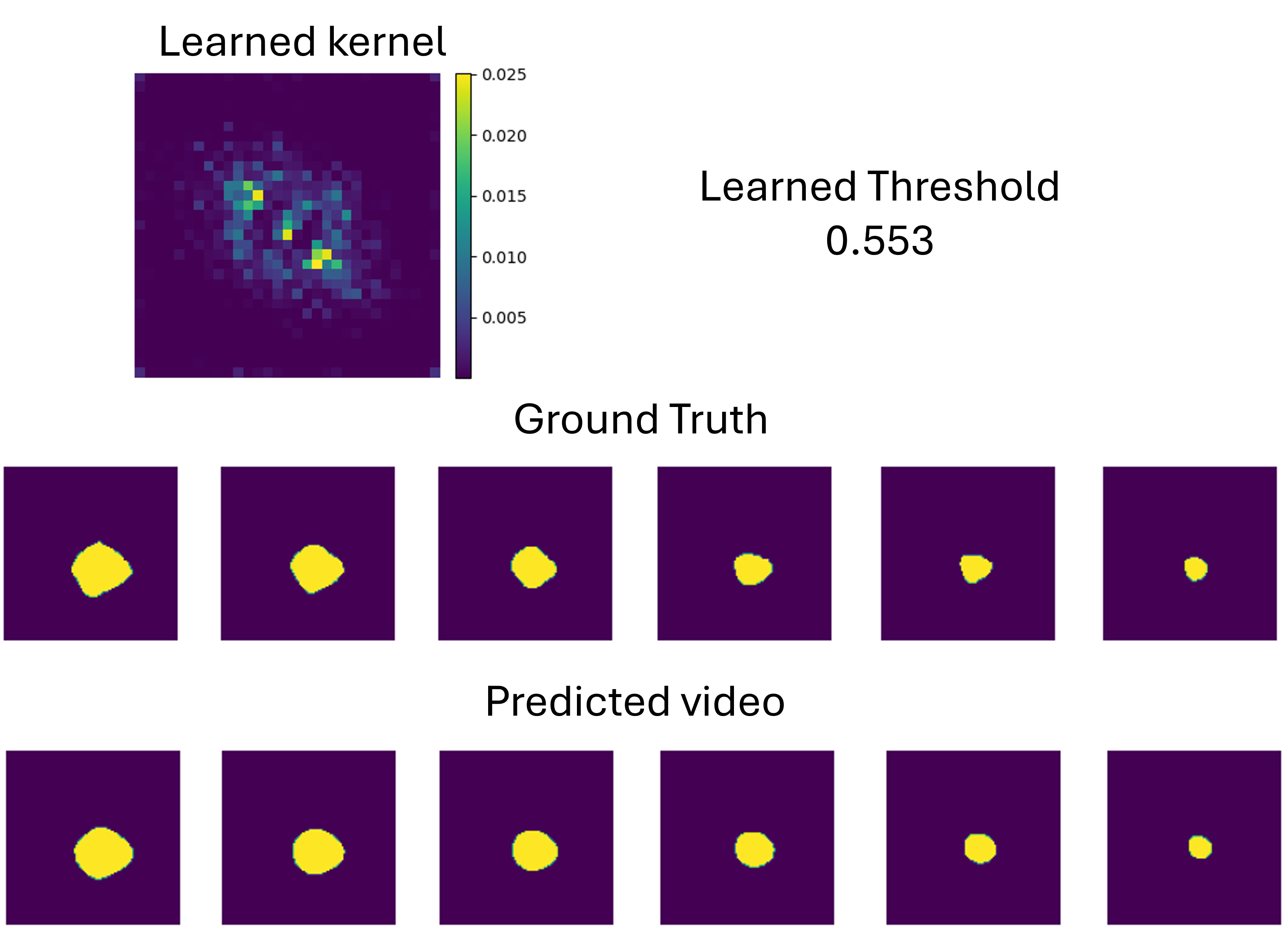}
    \caption{Results for one ice-melting test video when using meta-learning MBO trained on Gaussian blur data. \textbf{Top:} predicted kernel and threshold. \textbf{Bottom:} reconstructed and true dynamics.}
    \label{fig:pred_ice4}
\end{figure}

Figure \ref{fig:pred_ice4} shows the results of the meta-learning MBO network trained on Gaussian blur data on a specific test video. We can see that the reconstructed kernel resembles a skewed Gaussian kernel and the threshold is close to 0.5 as we expect from the theory. We can also see from the reconstructed video that the ice-melting dynamic is well captured by the model. However, some small differences between the reconstruction and the ground truth can be seen, for example the ground truth shows sharp edges that are not captured in the reconstruction. These differences are due to the nature of the real data, which does not strictly adhere to the theoretical dynamics. For instance, variations in ice thickness can lead to differing melting rates across the surface, which introduces variations in the dynamics that the model may not fully account for.

\textbf{Extrapolation.} In this section we test the ability of the two networks to extrapolate future dynamics. In this setting, given one video, we train one MBO-network using only the initial 4 frames to produce the kernel and threshold for that specific video. We repeat this procedure on each of the 5 videos in our dataset to get 5 trained MBO networks. 
For each trained network, we use the learned kernels and thresholds to predict the next three frames (frames 5, 6, and 7) and calculate the prediction error. By averaging these errors across the five networks, we obtain an overall measure of extrapolation accuracy, which assesses the MBO architecture’s ability to predict future dynamics on the ice-melting dataset. 
For the meta-learning MBO-network we use the same pre-trained architecture as in the previous section, but we only compute the prediction error on the last 3 frames of the ice-melting videos. Results are shown in Table \ref{tab:ice_extrap_metrics}.

\begin{table}[h!]
\centering
\caption{Performance comparison of the MBO network and the meta-learning MBO in the extrapolation of the last 3 frames of the ice-melting videos.}
\label{tab:ice_extrap_metrics}
\begin{tabular}{|c|c|c|c|}
\hline
\textbf{Model} & \textbf{Relative MSE $\downarrow$} & \textbf{SSIM Value $\uparrow$} & \textbf{Jaccard Index $\uparrow$} \\ \hline
\text{MBO Network} & 48.42\% & 0.917 & 61.15\% \\ \hline
\text{Meta-learning MBO (No Noise)} & 37.1\% & 0.926 & 70.1\% \\ \hline
\text{Meta-learning MBO (Gaussian Blur)} & \textbf{29.1\%} &  \textbf{0.930}& \textbf{73.3\%} \\ \hline
\end{tabular}

\end{table}
The MBO network results are improved compared the the previous section (especially SSIM, from 0.859 to 0.917, and Jaccard index from 52.4\% to 61.15\%). This is expected since we trained one MBO network per video. The meta-learning MBO results are generally worse or comparable to the previous case. This happens because the error in the prediction accumulates in time as past predicted frames are used to produce future ones. The best results for extrapolation are given by the meta-learning MBO network trained on blurry data.

\subsubsection{Fire Fronts Dataset}
\textbf{Generalization.} For the fire fronts dataset, we conducted a similar experiment to that of the ice-melting dataset. Using the MBO network, we trained on one fire video to learn the corresponding kernel and threshold, and used them to compute the model's prediction for the other 4 test videos. Given the challenges of real-world fire dynamics, we expect the models to perform less accurately on this dataset compared to the ice-melting data.

For the meta-learning MBO network, we again used the pre-trained model from Section \ref{metaL}, without any fine-tuning. Although the fire dynamics may differ from the synthetic data used in training, we expect the meta-learning model to approximate the kernel and threshold to a reasonable degree. We present results on the 4 test fire videos using both the noiseless and blurry pre-trained architectures for comparison. The results of these experiments can be found in Table \ref{tab:fire_recon_metrics}. In the case of the fire dataset, we see a similar trend as in the ice-melting example: the meta-learning MBO network trained with Gaussian blur produces the best results, with a relative MSE of 21.63\% and a Jaccard Index of 80.52\%, outperforming both the noiseless meta-learning MBO and the MBO network. As with the ice-melting data, this can be attributed to the fact that the real-world fire videos also inherently contain noise and variability that the blurred model can better account for during testing. Notably, the errors for fire videos are larger than for ice-melting, which aligns with expectations. Fire dynamics are harder to model due to the influence on our data of external, unpredictable factors like wind, weather, or physical barriers (such as lakes or mountains) that are not considered by the model.

\begin{table}[h!]
\centering
\caption{Performance comparison of the MBO network and the Meta-learning MBO in the reconstruction of fire front test videos.}
\label{tab:fire_recon_metrics}
\begin{tabular}{|c|c|c|c|}
\hline
\textbf{Model} & \textbf{Relative MSE $\downarrow$} & \textbf{SSIM Value $\uparrow$} & \textbf{Jaccard Index $\uparrow$} \\ \hline
\text{MBO Network} & 31.519\% & 0.536 & 73.429\% \\ \hline
\text{Meta-learning MBO (No Noise)} & 25.56\% & 0.736 & 75.60\% \\ \hline
\text{Meta-learning MBO (Gaussian Blur)} & \textbf{21.63\%} & \textbf{0.757} & \textbf{80.52\%} \\ \hline
\end{tabular}

\end{table} 

\begin{figure}
    \centering
    \includegraphics[width=0.8\linewidth]{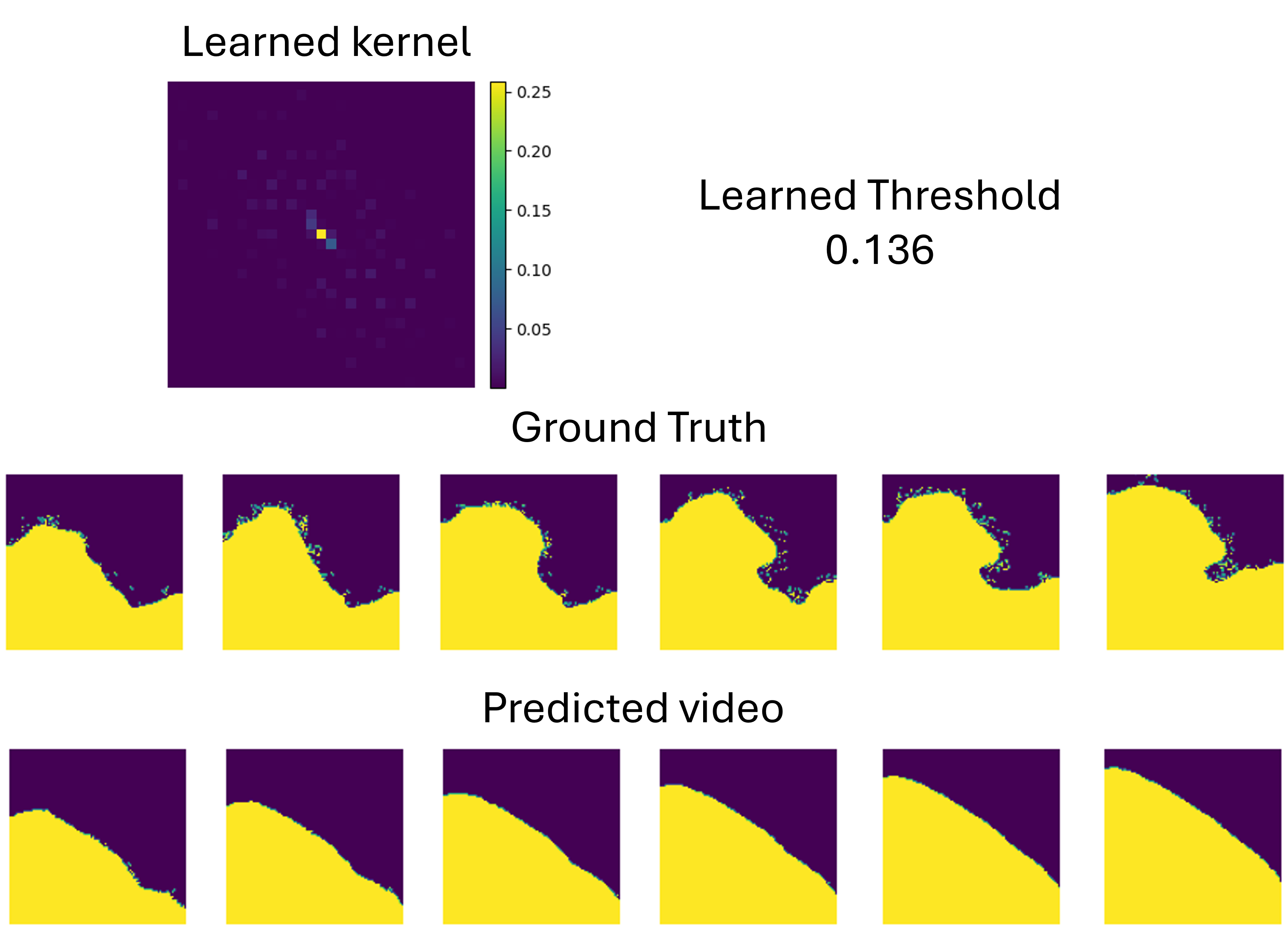}
    \caption{Results for one fire front test video when using meta-learning MBO trained on Gaussian blur data. \textbf{Top:} predicted kernel and threshold. \textbf{Bottom:} true and reconstructed dynamics.}
    \label{fig:pred_fire}
\end{figure}

Figure \ref{fig:pred_fire} shows the results of the meta-learning MBO network trained on Gaussian blur data on a test video. We can see that the reconstructed threshold is 0.136 which is, as expected from the theory, smaller than 0.5. The reconstructed kernel resembles the indicator function of a one-pixel disc. 
From the reconstructed video we can see that the fire dynamics are quite hard to reconstruct in contrast with the ice ones. The ground truth dynamic in this case does not evolve closely to the theoretical expected dynamics. In particular from the ground truth video we can see how different parts of the boundary expand at different speeds. This behavior is not accounted for in our model which justifies the discrepancy between the prediction and the ground truth. In particular, the true kernel for this observed dynamic may be more complicated than the indicator function of a disc or of any of the Gaussian and MNIST kernels used during training of the meta-learning MBO. In the video frames the yellow pixels represent the area of the ground that has been burnt or that contains an active fire. Note that, while our reconstruction cannot determine the boundary of such area exactly, it can still determine quite accurately an approximate area containing an active fire (lower left corner of the frame) which can still be useful in practice.

\textbf{Extrapolation.} In this section we analyze the ability of the network to extrapolate future fire fronts dynamics. Again, given one video, we use the first 4 frames to train the MBO network and produce a kernel and threshold for that video. We repeat this procedure on each of the 5 videos in our dataset and obtain 5 trained MBO networks. Finally, we use the learned kernels and thresholds for each video to produce the future 3 frames (frames 5, 6, 7) and compute the average error. For the meta-learning MBO network, we use the pre-trained architecture, but only compute the error for frames 5, 6, 7. The results are shown in Table \ref{tab:fire_extrap_metrics}.

\begin{table}[h!]
\centering
\caption{Performance comparison of the MBO network and the meta-learning MBO in the extrapolation of the last 3 frames of the fire front videos.}
\label{tab:fire_extrap_metrics}
\begin{tabular}{|c|c|c|c|}
\hline
\textbf{Model} & \textbf{Relative MSE $\downarrow$} & \textbf{SSIM Value $\uparrow$} & \textbf{Jaccard Index $\uparrow$} \\ \hline
\text{MBO Network} & \textbf{17.3\%} & \textbf{0.760} & \textbf{84.3\%} \\ \hline
\text{Meta-learning MBO (No Noise)} & 27.0\% & 0.701 & 74.1\% \\ \hline
\text{Meta-learning MBO (Gaussian Blur)} & 21.2\% &  0.751& 81.3\% \\ \hline
\end{tabular}

\end{table}

\begin{figure}
    \centering
    \includegraphics[width=0.8\linewidth]{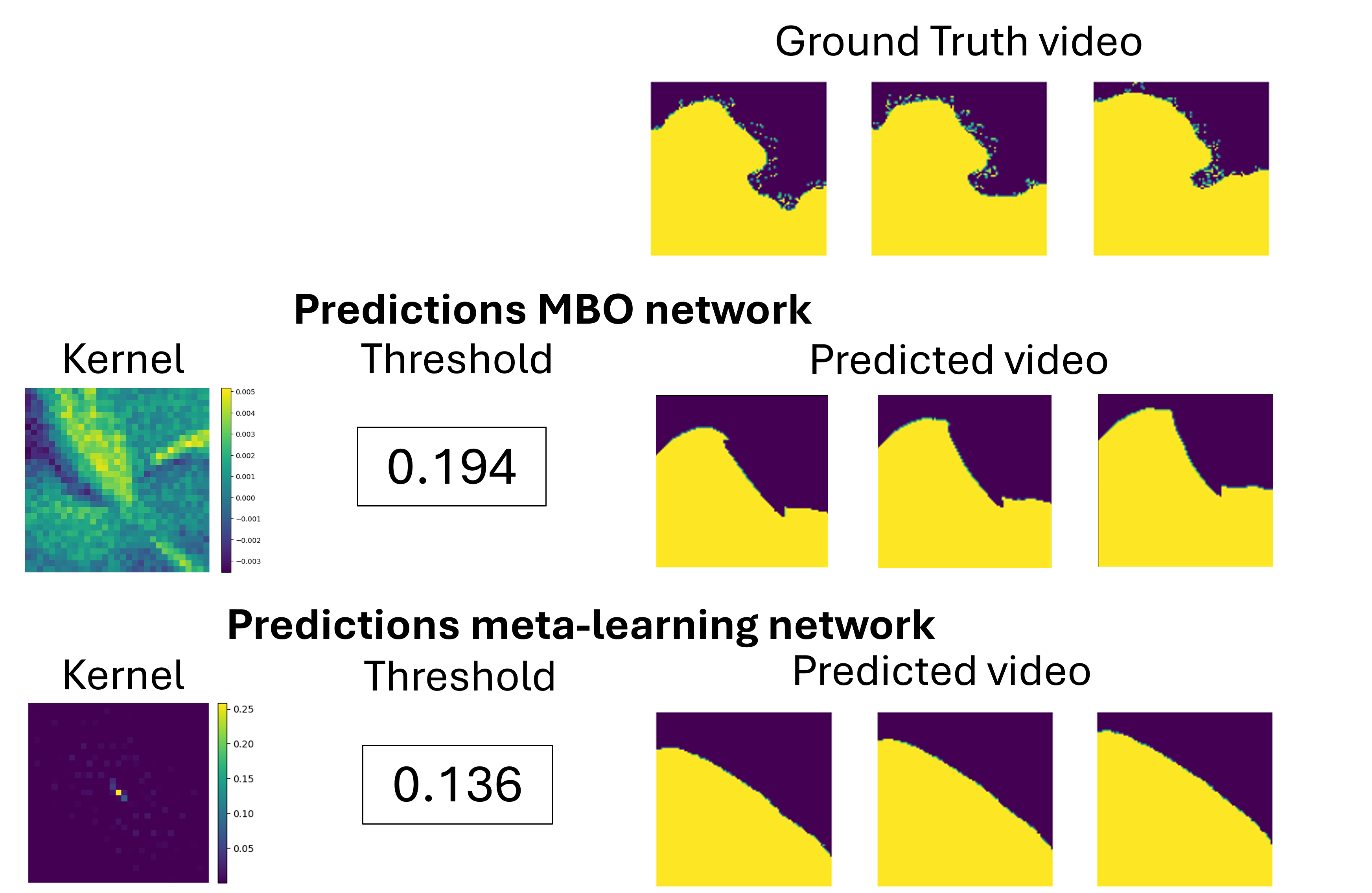}
    \caption{Comparison of extrapolation abilities of the MBO network and of meta-learning MBO on one fire front video. From left to right: predicted kernel, predicted threshold and predicted frames 4,5,6. From top to bottom: ground truth video, predictions by MBO network, predictions by meta-learning MBO network.}
    \label{fig:extrap_example}
\end{figure}

In this case the best results are obtained by the MBO network. A possible explanation is that in this case we train one MBO network per video, so each network can generate a specialized, possibly very complicated, kernel to match each video’s unique dynamics. In contrast, the meta-learning MBO network is pre-trained on synthetic kernels so it is constrained to produce kernels similar to those in its training set. This may be limiting its flexibility to adapt to the fire-front data. In fact, while the theoretical evolution of fire fronts is generated by disc-like kernels, accurate reconstruction of this dataset may require much more complicated kernels. An example of this can be seen in Figure \ref{fig:extrap_example} which compares reconstructed kernel, threshold and frames for one fire front video. The figure shows that the kernel reconstructed by the MBO network is much more complicated than the kernel estimated by the meta-learning MBO network and extremely different from the theoretical kernel (indicator function of a disc). 

We note explicitly that this limitation of the meta-learning architecture could be easily overcome by improving the training dataset, for example by adding real-data videos or fine-tuning for each specific real-data case. Since we only had access to very few real world videos we did not explore this direction and leave this for future work. 

From these experiments on real data we can draw a few conclusions on the best settings for each method:
\begin{itemize}
    \item For low-noise data (for example data collected in a controlled lab setup, like the ice-melting dataset) using the meta-learning MBO-network pre-trained on synthetic data may be the best choice as is can be directly and quickly used for predictions. This method will provide accurate results even on completely unseen dynamics as long as the kernels and threshold seen during training resemble the real data ones.
    \item For high-noise data (for example, the fire-fronts videos which were impacted by wind, physical barriers and noise that distorted the expected dynamics) the best choice may be to independently train one MBO network for each video and use it to produce future frames. While this may be more computationally expensive than using a pre-trained model, in practice training takes only a few seconds per network as the only learnable parameters are one kernel matrix (in the experiments 31x31 matrix) and one numeric threshold. Moreover, this specialized training allows the networks to produce complicated kernels which can possibly better capture noisy dynamics. Another possibility would be to fine-tune the meta-learning MBO model for the specific application.
\end{itemize}

\section{Conclusion and future work}
This work introduced two CNN architectures, the MBO network and the meta-learning MBO network, to learn threshold dynamics from video data. The MBO network excels in accurately capturing specific dynamics, while the meta-learning MBO network generalizes effectively across diverse and unseen dynamics. Experiments on synthetic and real-world datasets demonstrated the robustness and adaptability of both approaches. Future work could focus on exploring the following aspects:
\begin{enumerate}

\item {\bf Convergence.} 
We note a new theoretical question raised by our methods, which we hope to address in a future work. To the best of our knowledge, given the new thresholding scheme \eqref{new_scheme} based on a sigmoid thresholding function it is an open question whether the convergence
$$\tilde u_{\frac{t}{m}}^m \to \mathcal M (t) u^0$$
as $m \rightarrow +\infty $ holds, where $\mathcal M(t) u^0$ denotes a viscosity solution to the mean curvature flow PDE \eqref{MCF}.
\item {\bf Improving interpretability of the network by learning the PDE.} In the paper \cite{IPS}, Ishii, Pires and Souganidis prove that the threshold dynamics schemes studied in the present paper converge to the evolution of level sets of weak solutions $u(x,t)$ of a degenerate parabolic, fully nonlinear PDE of the form
$$\partial_t u = |\nabla u| \Bigg(\text{Trace} \bigg[E(\vec n) D\vec n\bigg]+\nu(\vec n)\Bigg),$$
where $\vec n=\nabla u/|\nabla u|$ denotes the unit normal to the level sets, and the functions $E, \nu$ can be recovered from the convolution kernel and the threshold associated to the thresholding scheme. In a future work we plan to implement this derivation and recover the threshold dynamics limiting PDE from data. 
\item {\bf Esodoglu-Otto (multiple junctions)} In the paper \cite{MR3333842} Esedoglu and Otto introduce a new algorithm to approximate the  mean curvature motion for an arbitrary set of (isotropic) surface tensions. This algorithm is a variant of the MBO scheme and is based on minimizing interfacial energy. In a future work we aim at implementing this energy minimization approach through loss functions.

\end{enumerate}

\printcredits

\bibliographystyle{cas-model2-names}

\bibliography{reference.bib}

\begin{thebibliography}{22}
\expandafter\ifx\csname natexlab\endcsname\relax\def\natexlab#1{#1}\fi
\providecommand{\url}[1]{\texttt{#1}}
\providecommand{\href}[2]{#2}
\providecommand{\path}[1]{#1}
\providecommand{\DOIprefix}{doi:}
\providecommand{\ArXivprefix}{arXiv:}
\providecommand{\URLprefix}{URL: }
\providecommand{\Pubmedprefix}{pmid:}
\providecommand{\doi}[1]{\href{http://dx.doi.org/#1}{\path{#1}}}
\providecommand{\Pubmed}[1]{\href{pmid:#1}{\path{#1}}}
\providecommand{\bibinfo}[2]{#2}
\ifx\xfnm\relax \def\xfnm[#1]{\unskip,\space#1}\fi
\bibitem[{Aeronautics and (NASA)()}]{firms}
\bibinfo{author}{Aeronautics, N.}, \bibinfo{author}{(NASA), S.A.}, .
\newblock \bibinfo{title}{Fire information for resource management system (firms)}.
\newblock \bibinfo{howpublished}{NASA Earth Observing System Data and Information System (EOSDIS)}.
\newblock \URLprefix \url{https://firms.modaps.eosdis.nasa.gov/}. \bibinfo{note}{access to FIRMS data requires acknowledgement of the LANCE Citation, Acknowledgements, and Disclaimer}.
\bibitem[{Barles(1985)}]{Ba}
\bibinfo{author}{Barles, G.}, \bibinfo{year}{1985}.
\newblock \bibinfo{title}{Remarks on a flame propagation model}.
\newblock Ph.D. thesis. INRIA.
\bibitem[{Barles and Georgelin(1995)}]{BG}
\bibinfo{author}{Barles, G.}, \bibinfo{author}{Georgelin, C.}, \bibinfo{year}{1995}.
\newblock \bibinfo{title}{A simple proof of convergence for an approximation scheme for computing motions by mean curvature}.
\newblock \bibinfo{journal}{SIAM Journal on Numerical Analysis} \bibinfo{volume}{32}, \bibinfo{pages}{484--500}.
\bibitem[{Bertozzi and Flenner(2012)}]{bertozzi2012diffuse}
\bibinfo{author}{Bertozzi, A.L.}, \bibinfo{author}{Flenner, A.}, \bibinfo{year}{2012}.
\newblock \bibinfo{title}{Diffuse interface models on graphs for classification of high dimensional data}.
\newblock \bibinfo{journal}{Multiscale Modeling \& Simulation} \bibinfo{volume}{10}, \bibinfo{pages}{1090--1118}.
\bibitem[{Calder et~al.(2020)Calder, Cook, Thorpe and Slepcev}]{calder2020poisson}
\bibinfo{author}{Calder, J.}, \bibinfo{author}{Cook, B.}, \bibinfo{author}{Thorpe, M.}, \bibinfo{author}{Slepcev, D.}, \bibinfo{year}{2020}.
\newblock \bibinfo{title}{Poisson learning: Graph based semi-supervised learning at very low label rates}, in: \bibinfo{booktitle}{International Conference on Machine Learning}, \bibinfo{organization}{PMLR}. pp. \bibinfo{pages}{1306--1316}.
\bibitem[{Chen et~al.(1999)Chen, Giga and Goto}]{CGG}
\bibinfo{author}{Chen, Y.G.}, \bibinfo{author}{Giga, Y.}, \bibinfo{author}{Goto, S.}, \bibinfo{year}{1999}.
\newblock \bibinfo{title}{Uniqueness and existence of viscosity solutions of generalized mean curvature flow equations}, in: \bibinfo{booktitle}{Fundamental Contributions to the Continuum Theory of Evolving Phase Interfaces in Solids: A Collection of Reprints of 14 Seminal Papers}. \bibinfo{publisher}{Springer}, pp. \bibinfo{pages}{375--412}.
\bibitem[{Esedo\=glu and Otto(2015)}]{MR3333842}
\bibinfo{author}{Esedo\=glu, S.}, \bibinfo{author}{Otto, F.}, \bibinfo{year}{2015}.
\newblock \bibinfo{title}{Threshold dynamics for networks with arbitrary surface tensions}.
\newblock \bibinfo{journal}{Comm. Pure Appl. Math.} \bibinfo{volume}{68}, \bibinfo{pages}{808--864}.
\newblock \URLprefix \url{https://doi.org/10.1002/cpa.21527}, \DOIprefix\doi{10.1002/cpa.21527}.
\bibitem[{Esedo\=glu et~al.(2005)Esedo\=glu, Ruuth and Tsai}]{ERT}
\bibinfo{author}{Esedo\=glu, S.}, \bibinfo{author}{Ruuth, S.}, \bibinfo{author}{Tsai, R.}, \bibinfo{year}{2005}.
\newblock \bibinfo{title}{Threshold dynamics for shape reconstruction and disocclusion}, in: \bibinfo{booktitle}{IEEE International Conference on Image Processing 2005}, pp. \bibinfo{pages}{II--502}.
\newblock \DOIprefix\doi{10.1109/ICIP.2005.1530102}.
\bibitem[{Esedo\=glu and Tsai(2006)}]{ESEDOGLU2006367}
\bibinfo{author}{Esedo\=glu, S.}, \bibinfo{author}{Tsai, Y.H.R.}, \bibinfo{year}{2006}.
\newblock \bibinfo{title}{Threshold dynamics for the piecewise constant mumford–shah functional}.
\newblock \bibinfo{journal}{Journal of Computational Physics} \bibinfo{volume}{211}, \bibinfo{pages}{367--384}.
\newblock \URLprefix \url{https://www.sciencedirect.com/science/article/pii/S0021999105002792}, \DOIprefix\doi{https://doi.org/10.1016/j.jcp.2005.05.027}.
\bibitem[{Evans(1993)}]{E}
\bibinfo{author}{Evans, L.C.}, \bibinfo{year}{1993}.
\newblock \bibinfo{title}{Convergence of an algorithm for mean curvature motion}.
\newblock \bibinfo{journal}{Indiana University mathematics journal} , \bibinfo{pages}{533--557}.
\bibitem[{Evans and Spruck(1991)}]{ES}
\bibinfo{author}{Evans, L.C.}, \bibinfo{author}{Spruck, J.}, \bibinfo{year}{1991}.
\newblock \bibinfo{title}{Motion of level sets by mean curvature. i}, in: \bibinfo{booktitle}{Fundamental Contributions to the Continuum Theory of Evolving Phase Interfaces in Solids: A Collection of Reprints of 14 Seminal Papers}. \bibinfo{publisher}{Springer}, pp. \bibinfo{pages}{328--374}.
\bibitem[{Garcia-Cardona et~al.(2014)Garcia-Cardona, Merkurjev, Bertozzi, Flenner and Percus}]{garcia2014multiclass}
\bibinfo{author}{Garcia-Cardona, C.}, \bibinfo{author}{Merkurjev, E.}, \bibinfo{author}{Bertozzi, A.L.}, \bibinfo{author}{Flenner, A.}, \bibinfo{author}{Percus, A.G.}, \bibinfo{year}{2014}.
\newblock \bibinfo{title}{Multiclass data segmentation using diffuse interface methods on graphs}.
\newblock \bibinfo{journal}{IEEE transactions on pattern analysis and machine intelligence} \bibinfo{volume}{36}, \bibinfo{pages}{1600--1613}.
\bibitem[{Gravner and Griffeath(1993)}]{GrGr}
\bibinfo{author}{Gravner, J.}, \bibinfo{author}{Griffeath, D.}, \bibinfo{year}{1993}.
\newblock \bibinfo{title}{Threshold growth dynamics}.
\newblock \bibinfo{journal}{Transactions of the American Mathematical society} , \bibinfo{pages}{837--870}.
\bibitem[{Ishii et~al.(1999)Ishii, Pires and Souganidis}]{IPS}
\bibinfo{author}{Ishii, H.}, \bibinfo{author}{Pires, G.E.}, \bibinfo{author}{Souganidis, P.E.}, \bibinfo{year}{1999}.
\newblock \bibinfo{title}{Threshold dynamics type approximation schemes for propagating fronts}.
\newblock \bibinfo{journal}{Journal of the Mathematical Society of Japan} \bibinfo{volume}{51}, \bibinfo{pages}{267--308}.
\bibitem[{Jacobs et~al.(2018)Jacobs, Merkurjev and Esedoḡlu}]{jacobs2018auction}
\bibinfo{author}{Jacobs, M.}, \bibinfo{author}{Merkurjev, E.}, \bibinfo{author}{Esedoḡlu, S.}, \bibinfo{year}{2018}.
\newblock \bibinfo{title}{Auction dynamics: A volume constrained mbo scheme}.
\newblock \bibinfo{journal}{Journal of Computational Physics} \bibinfo{volume}{354}, \bibinfo{pages}{288--310}.
\bibitem[{{Jona Lelmi}(2023)}]{handle:20.500.11811/10889}
\bibinfo{author}{{Jona Lelmi}}, \bibinfo{year}{2023}.
\newblock \bibinfo{title}{Analysis of the {MBO} scheme: from materials science to data clustering}.
\newblock Ph.D. thesis. Rheinische Friedrich-Wilhelms-Universität Bonn.
\newblock \URLprefix \url{https://hdl.handle.net/20.500.11811/10889}.
\bibitem[{Laux and Lelmi(2023)}]{JMLR:v24:22-1089}
\bibinfo{author}{Laux, T.}, \bibinfo{author}{Lelmi, J.}, \bibinfo{year}{2023}.
\newblock \bibinfo{title}{Large data limit of the mbo scheme for data clustering: convergence of the dynamics}.
\newblock \bibinfo{journal}{Journal of Machine Learning Research} \bibinfo{volume}{24}, \bibinfo{pages}{1--49}.
\newblock \URLprefix \url{http://jmlr.org/papers/v24/22-1089.html}.
\bibitem[{Mascarenhas(1992)}]{Ma}
\bibinfo{author}{Mascarenhas, P.}, \bibinfo{year}{1992}.
\newblock \bibinfo{title}{Diffusion generated motion by mean curvature}.
\newblock \bibinfo{publisher}{CAM Reports, Department of Mathematics, University of California, Los Angeles}.
\bibitem[{Merkurjev et~al.(2014)Merkurjev, Sunu and Bertozzi}]{merkurjev2014graph}
\bibinfo{author}{Merkurjev, E.}, \bibinfo{author}{Sunu, J.}, \bibinfo{author}{Bertozzi, A.L.}, \bibinfo{year}{2014}.
\newblock \bibinfo{title}{Graph mbo method for multiclass segmentation of hyperspectral stand-off detection video}, in: \bibinfo{booktitle}{2014 IEEE International Conference on Image Processing (ICIP)}, \bibinfo{organization}{IEEE}. pp. \bibinfo{pages}{689--693}.
\bibitem[{Merriman et~al.(1992)Merriman, Bence and Osher}]{MBO}
\bibinfo{author}{Merriman, B.}, \bibinfo{author}{Bence, J.K.}, \bibinfo{author}{Osher, S.}, \bibinfo{year}{1992}.
\newblock \bibinfo{title}{Diffusion generated motion by mean curvature}.
\newblock \bibinfo{publisher}{CAM Reports, Department of Mathematics, University of California, Los Angeles}.
\bibitem[{Merriman et~al.(1994)Merriman, Bence and Osher}]{MBO1}
\bibinfo{author}{Merriman, B.}, \bibinfo{author}{Bence, J.K.}, \bibinfo{author}{Osher, S.J.}, \bibinfo{year}{1994}.
\newblock \bibinfo{title}{Motion of multiple junctions: A level set approach}.
\newblock \bibinfo{journal}{Journal of computational physics} \bibinfo{volume}{112}, \bibinfo{pages}{334--363}.
\bibitem[{Sethian(1982)}]{Se}
\bibinfo{author}{Sethian, J.A.}, \bibinfo{year}{1982}.
\newblock \bibinfo{title}{An analysis of flame propagation}.
\newblock \bibinfo{publisher}{University of California, Berkeley}.

\end{thebibliography}
\newpage
\appendix
\section{Appendix}
\subsection{MBO network: learning with one short video}\label{one_short_app}
\begin{figure}[h!]
    \centering
    \includegraphics[width=0.8\linewidth]{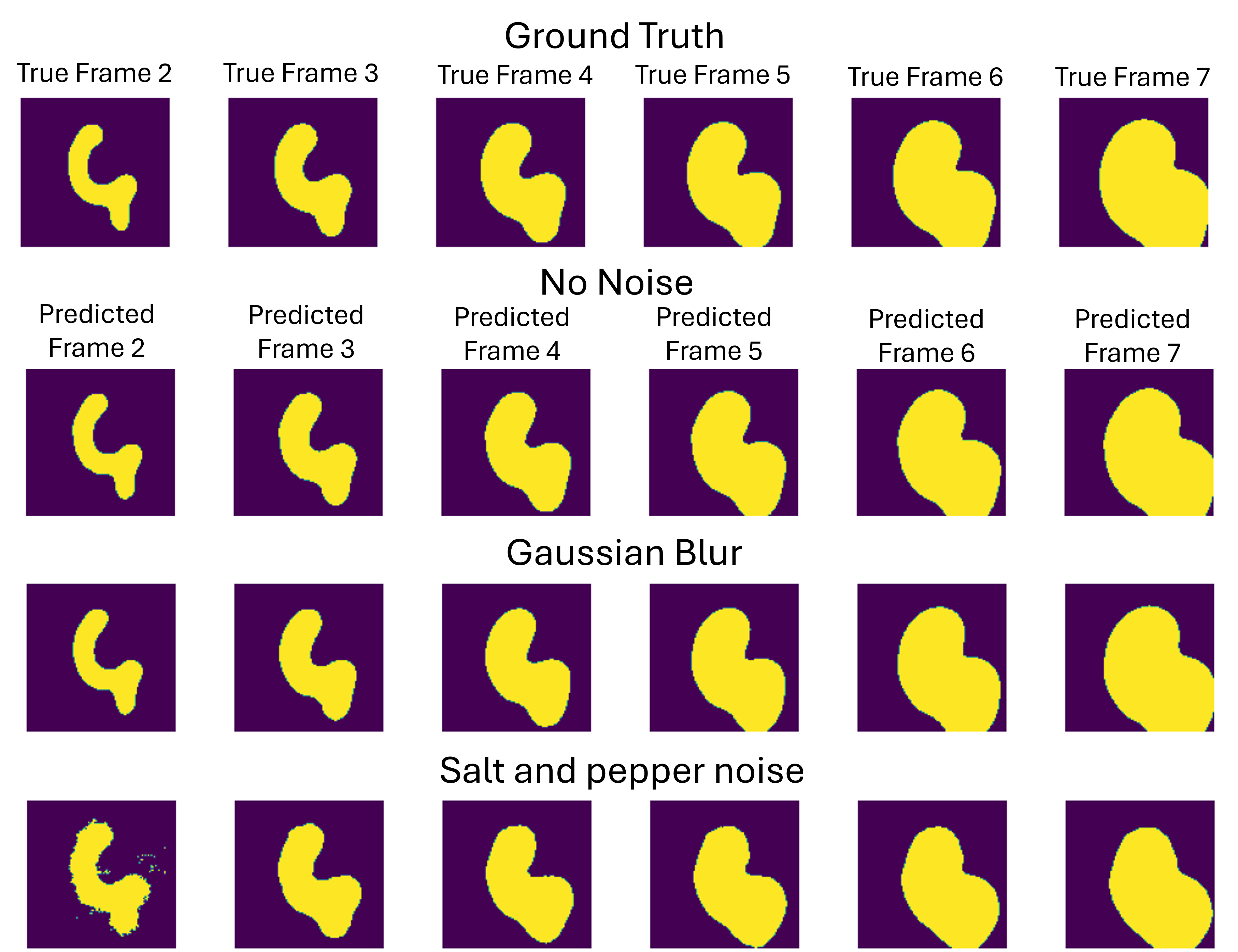}
    \caption{Reconstructed and true test video for standard Gaussian kernel and threshold 0.2. From left to right we show frames 1 through 7. From top to bottom we show the ground truth video and the reconstructed video respectively in the case of no noise, Gaussian blur and salt-and-pepper noise. Training was done using only one 4 frames video.}
    \label{fig:1vidSG_recon}
\end{figure}

We show here a test video reconstruction for the case of training the MBO network with one short video. The corresponding kernel and threshold are shown in Section \ref{sec:one_vid}, Figure \ref{fig:1vidSGkernel_recon}. The effects of using less data for training (in this case only one video) are visible in Figure \ref{fig:1vidSG_recon}: accurate results are obtained for no noise and blur conditions, while the salt-and-pepper noise case exhibits an incorrect support for the digit, especially in the later predicted frames.

\end{document}